%% file: paper.tex
\documentclass[11pt]{article}
\usepackage{hyperref}
\usepackage[ruled, lined]{algorithm2e}

\usepackage{amssymb}
\include{macros}

 \newcommand{\myvecC}{\{\myvec{x} \mid ~C ~\And~\myvec{x}\in\myvec{X}\}}

 \newcommand{\XvecXmax}{\{\myvec{x} \mid ~x_i=max(X_i) ~\And~\myvec{x}\in\myvec{X}\}}
 \newcommand{\minXvecXmax}{min\{\myvec{x} \mid~x_i=max(X_i) ~\And~\myvec{x}\in\myvec{X}\}}
 \newcommand{\YvecYmin}{\{\myvec{y} \mid ~y_i=min(Y_i) ~\And~\myvec{y}\in\myvec{Y}\}}
 \newcommand{\maxYvecYmin}{max\{\myvec{y} \mid ~y_i=min(Y_i) ~\And~\myvec{y}\in\myvec{Y}\}}

\newcommand{\mset}{\mathbf}
\newcommand{\msetl}{\{ \hspace{-0.3em} \{}
\newcommand{\msetr}{\} \hspace{-0.3em} \}}

\newcommand{\myvec}{\vec}
\newcommand{\domain}{\mathcal}
\newcommand{\mydomain}[1]{\ensuremath{\mathcal #1}}
\newcommand{\groundandequal}{\doteq}

\newcommand{\assigned}{\leftarrow}

\newcommand{\myalldiff}{\mbox{\it all-different}}

\newcommand{\myfloor}{{\tt floor}}
\newcommand{\myceiling}{{\tt ceiling}}

\newcommand{\MsetLeq}{\tt{MsetLeq}}
\newcommand{\MsetLess}{\tt{MsetLess}}

\newtheorem{proposition}{Proposition}
\newtheorem{theorem}{Theorem}

\newtheorem{definition}{Definition}

\newcommand{\proof}{\noindent {\bf Proof:\ \ }}
\newcommand{\qed}{\mbox{QED.}}

\usepackage{ifthen}
\newlength{\margintonumber}
\newcommand{\numline}[2]{
   \ifthenelse{#2>-1}{
     \setlength{\margintonumber}{#2\algomargin}
     \addtolength{\margintonumber}{2em}
     \hspace{-\margintonumber}
     {\textsf{\footnotesize\raggedleft\textbf{#1}}}
     \settowidth{\margintonumber}{\textsf{\footnotesize\raggedleft\textbf{#1}}}
     \addtolength{\margintonumber}{.427in}
     \hspace{-\margintonumber}
     \setlength{\margintonumber}{#2\algomargin}
     \addtolength{\margintonumber}{2em}
     \hspace{\margintonumber}
   }
   { 
     \typeout{^^JError using command \\numline (argument #2 should be
positive)^^J}
   }
   }

\begin{document}

\title{\textbf{Filtering Algorithms \\ for the Multiset Ordering Constraint}}


\author{Alan M. Frisch \\ Department of Computer Science, University of York, UK. \\ {\tt
frisch@cs.york.ac.uk} \and Brahim Hnich  \\ Faculty of Computer
Science, Izmir University of Economics, Turkey. \\ {\tt
brahim.hnich@ieu.edu.tr} \and Zeynep Kiziltan \\ Department of
Computer Science, University of Bologna, Italy. \\ {\tt
zeynep@cs.unibo.it} \and Ian Miguel \\ School of Computer Science,
University of St Andrews, UK. \\ {\tt ianm@dcs.st-and.ac.uk} \and
Toby Walsh \\ NICTA and School of Computer Science
and Engineering, \\ University of New South Wales, Australia.  {\tt
toby.walsh@nicta.com.au }  }

\maketitle
\date

\begin{abstract}

\noindent Constraint programming (CP) has been used with great
success to tackle a wide variety of constraint satisfaction problems
which are computationally intractable in general. Global constraints
are one of the important factors behind the success of CP. In this
paper, we study a new global constraint, the multiset
ordering constraint, which is shown to be useful in symmetry
breaking and searching for leximin optimal solutions in CP.
We propose efficient and effective filtering algorithms for
propagating this global constraint. We show that the algorithms
maintain generalised arc-consistency and we discuss possible
extensions. We also consider alternative propagation methods based
on existing constraints in CP toolkits. Our experimental results on
a number
of benchmark problems demonstrate 
that propagating the multiset ordering constraint via a dedicated
algorithm can be very beneficial.
\end{abstract}


\section{Introduction}
\label{sec:Intro}



Constraint satisfaction problems (CSPs) play an important role in
various fields of computer science \cite{tsang:book93} and are
ubiquitous in many real-life application areas such as production
planning, staff scheduling, resource allocation, circuit design,
option trading, and DNA sequencing. In general, solving CSPs is
NP-hard and so is computationally intractable \cite{Mackworth77a}.
{\em Constraint programming}
 (CP) provides a platform for solving CSPs
 \cite{CPbook:ms}\cite{CPbook:apt}
 and has proven successful in many real-life applications
\cite{wallace:practicalCP}\cite{rossi:survey}\cite{CPhandbook}
despite this intractability.
One of the jewels of CP is the notion of global (or non-binary)
constraints.
They encapsulate patterns that occur frequently in constraint
models. Moreover, they contain specialised filtering algorithms for
powerful constraint inference. Dedicated filtering algorithms for
global constraints are vital for efficient and effective constraint
solving. A number of such algorithms have been developed
(see \cite{Beldiceanu:catalog} for examples).


In this paper, we study a new global constraint, the
multiset ordering constraint, which ensures that the values taken by
two vectors of variables, when viewed as multisets, are ordered.
This constraint has applications in breaking row and column symmetry
as well as in searching for leximin optimal solutions. We
propose two different filtering algorithms for the multiset ordering
(global) constraint. Whilst they both maintain generalised
arc-consisteny, they differ in their complexity.
The first algorithm {\MsetLeq}  runs in time that is in the number of
variables ($n$) and in the
number of distinct values ($d$)
and is suitable when $n$ is much bigger than $d$.
Instead, the second algorithm is
more suitable when we have large domains and runs in time
$O(nlog(n))$ independent of $d$. We propose further algorithms by considering
some extensions to {\MsetLeq}. In particular, we show how we
can identify entailment and obtain a filtering algorithm for the strict
multiset ordering constraint.
These algorithms are proven to maintain generalised arc-consistency.

We consider alternative approaches to propagating the multiset
ordering constraint by using existing constraints in CP toolkits. We
evaluate our algorithms in contrast to the alternative approaches on
a variety of representative problems in the context of symmetry
breaking. The results demonstrate that our filtering algorithms are
superior to the alternative approaches either in terms of pruning
capabilities or in terms of computational times or both. We stress
that the contribution of this paper is the study of the filtering
algorithms for the multiset ordering constraint. Symmetry breaking
is merely used to compare the efficiency of these propagators. A
more in depth comparison of symmetry breaking methods awaits a
separate study. Such a study would be interesting in its own right
as multiset ordering constraints are one of the few methods for
breaking symmetry which are not special cases of lexicographical
ordering constraints \cite{clgrkr96}. Nevertheless, we provide
experimental evidence to support the need of multiset ordering
consraints in the context of symmetry breaking.

The rest of the paper is organised as follows. After we give the
necessary formal background in the next section, we present in
Section \ref{sec:App} the utility of the multiset ordering
constraint. In Section \ref{ch-gacmset:algorithm}, we present our
first filtering
algorithm
, prove that it maintains generalised arc-consistency, and discuss
its complexity. Our second algorithm is introduced in Section
\ref{ch-gacmset:algo2}. In Section \ref{ch-gacmset:extensions}, we
extend our first algorithm to obtain an algorithm for the strict
multiset ordering constraint and to detect
entailment. 
Alternative propagation methods are discussed in Section
\ref{ch-gacmset:alternativeapproaches}. We demonstrate in Section
\ref{ch-gacmset:multiple vectors} that decomposing a chain of
multiset ordering constraints into multiset ordering constraints
between adjacent or all pairs of vectors hinders constraint
propagation. Computational results are presented in Section
\ref{ch-gacmset:experiments}. Finally, we conclude and outline our
plans for future work in Section \ref{ch-gacmset:summary}.

\section{Formal Background}
\label{sec:back}

\subsection{Constraint Satisfaction Problems And Constraint Programming}
\label{sec:csp}

A {\em finite-domain constraint satisfaction problem} (CSP) consists
of: (i) a finite set of variables $\domain{X}$; (ii) for each
variable $X \in \domain{X}$, a finite set $\domain{D}(X)$ of values
(its domain); (iii)  and a finite set $\domain{C}$ of constraints on
the variables, where each constraint $c(X_i,\ldots,X_j) \in
\domain{C}$ is defined over
the variables $X_i, \ldots, X_j$ by a subset of 
$\domain{D}(X_i) \times \cdots \times \domain{D}(X_j)$ giving the
set of allowed combinations of values.  That is, $c$ is an $n$-ary
relation.

A variable {\em assignment} or {\em instantiation} is an assignment
to a variable $X$ of one of the values from $\domain{D}(X)$. Whilst
a {\em partial assignment} $A$ to $\domain{X}$ is an assignment to
some but not all $X \in \domain{X}$, a {\em total
assignment}\footnote{Throughout, we will say {\em assignment} when
we mean {\em total assignment} to the problem variables.} $A$ to
$\domain{X}$ is an assignment to every $X \in \domain{X}$. We use
the notation $A[\domain{S}]$ to denote the projection of $A$ on to
the set of variables $\domain{S}$. A (partial) assignment $A$ to the
set of variables $\domain{T}\subseteq \domain{X}$ is {\em
consistent} iff for all constraints $c(X_i,\ldots,X_j) \in
\domain{C}$ such that $\{X_i, \ldots,X_j\} \subseteq \domain{T}$, we
have $A[\{X_i, \ldots,X_j\}] \in c(X_i,\ldots,X_j)$. A {\em
solution} to the CSP is a consistent assignment to $\domain{X}$.
A CSP is said to be satisfiable if it has a solution; otherwise it
is unsatisfiable. Typically, we are interested in finding one or all
solutions, or an optimal solution given some objective function. In
the presence of an objective function, a CSP is a {\em constraint
optimisation problem}.

Constraint Programming (CP) has been used with great success to
solve CSPs. Recent years have witnessed the development of several
CP systems \cite{CPhandbook}. To solve a problem using CP, we need
first to formulate it as a CSP by declaring the variables, their
domains, as well as the constraints on the variables. This part of
the problem solving is called {\em modelling}. In the following, we
first introduce our notations and then briefly overview modelling
and solving in CP. Since we compare our algorithms against the
alternative approaches in the context of symmetry breaking, we also
briefly review matrix modelling and index symmetry.

\subsection{Notation}

Throughout, we assume finite integer domains, which are totally
ordered. The domain of a variable $X$
is denoted by $\domain{D}(X)$, and the minimum and the maximum
elements in this domain by $min(X)$ and $max(X)$. We use {\em
vars}$(c)$  to denote the set of variables constrained by constraint
$c$. If a variable $X$ has a singleton domain $\{v\}$ we say that
$v$ is assigned to $X$ and denotes this by $X \assigned v$, or
simply say that $X$ is assigned. If two variables $X$ and $X'$ are
assigned the same value, then we write $X \groundandequal X'$,
otherwise we write $\Not(X \groundandequal X')$.

A one-dimensional matrix, or vector, is an ordered list of elements.
We denote a vector of $n$ variables as $\myvec X=\la
X_0,\ldots,X_{n-1}\ra$ and a vector of $n$ integers as $\myvec x=\la
x_0,\ldots,x_{n-1}\ra$. In either case, a sub-vector from index $a$
to index $b$ inclusive is denoted by the subscript $a \rightarrow
b$, such as: $\myvec x_{a \rightarrow b}$. Unless otherwise stated,
the indexing of vectors is from left to right, with $0$ being the
most significant index, and the variables of a vector $\myvec{X}$
are assumed to be disjoint and not repeated. The vector $\myvec
X_{X_i \assigned d}$ is the vector $\myvec X$ with some $X_i$ being
assigned to $d$. The functions \myfloor($\myvec X$) and
\myceiling($\myvec X$) assign all the variables of $\myvec{X}$ their
minimum and maximum values, respectively. A vector $\myvec{x}$ in
the domain of $\myvec{X}$ is designated by $\myvec{x} \in
\myvec{X}$. We write $\myvecC$ to denote the set of vectors in the
domain of $\myvec{X}$ which satisfy condition $C$. A vector of
variables is displayed by a vector of the domains of the
corresponding variables. For instance,  $ \myvec X
 = \la \{1,3,4\},  \{1,2,3,4,5\},  \{1,2\}\} \ra$ denotes the
vector of three variables whose domains are $\{1,3,4\}$, $
\{1,2,3,4,5\}$, and $ \{1,2\}$, respectively.


A set\index{Set} is an unordered list of elements in which
repetition is not allowed.  We denote a set of $n$ elements as
$\domain{X}=\{ x_0, \ldots, x_{n-1} \}$. A multiset\index{Multiset}
is an unordered list of elements in which repetition is allowed. We
denote a multiset of $n$ elements as $\mset{x}=\msetl x_0, \ldots,
x_{n-1} \msetr $. We write $max(\mset{x})$ or $max\msetl x_0,
\ldots, x_{n-1} \msetr$ for the maximum element of a multiset
$\mset{x}$. By ignoring the order of elements in a vector, we can
view a vector as a multiset. For example, the vector $\la 0, 1, 0
\ra$ can be viewed as the multiset $\msetl 1, 0, 0\msetr $. We will
abuse notation and write $\msetl \myvec{x} \msetr$ or $\msetl \la
x_0, \ldots, x_{n-1} \ra \msetr$ for the multiset view of the vector
$\myvec{x}=\la x_0, \ldots, x_{n-1} \ra$.

An occurrence vector $occ(\myvec x)$ associated with $\myvec x$ is
indexed in decreasing order of significance from the maximum
$max\msetl \myvec{x} \msetr$ to the minimum $min\msetl \myvec{x}
\msetr$ value from the values in $\msetl \myvec x \msetr$. The $i$th
element of $occ(\myvec x)$ is the number of occurrences of
$max\msetl \myvec{x} \msetr-i$ in $\msetl \myvec x \msetr$. When
comparing two occurrence vectors, we assume they start and end with
the occurrence of the same value, adding leading/trailing zeroes as
necessary. Finally, $sort(\myvec{x})$ is the vector obtained by
sorting the values in $\myvec{x}$ in non-increasing order.

\subsection{Search, Local Consistency and Propagation}

Solutions to CSPs are often found by {\em searching}
systematically the space of partial assignments. A common search
strategy is {\em backtracking search}. We traverse the search space
in a depth-first manner and at each step extend a partial
assignment by assigning a value to one more variable. If the
extended assignment is consistent then one more variable is
instantiated and so on. Otherwise, the variable is re-instantiated
with another value. If none of the values in the domain of the
variable is consistent with the current partial assignment then one
of the previous variable assignments is reconsidered.

Backtracking search may be seen as a {\em search tree} traversal.
Each node defines a partial assignment and each branch defines a
variable assignment. A partial assignment is extended by branching
from the corresponding node to one of its subtrees by assigning a
value $j$ to the next variable $X_i$ from the current
$\domain{D}(X_i)$. Upon backtracking, $j$ is removed from
$\domain{D}(X_i)$. This process is often called {\em labelling}. The
order of the variables and values chosen for consideration can have
a profound effect on the size of the search tree \cite{FC}. The
order can be determined before search starts, in which case the
labelling heuristic is {\em static}. If the next variable and/or
value are determined during search then the labelling heuristic is
{\em dynamic}.

The size of the search tree of a CSP is in the worst
case equal to the product of the domain
sizes of all variables. It is thus too expensive in general to enumerate
all possible assignments using a naive backtracking algorithm.
Consequently, many CP solution methods are based on {\em inference}
which reduces the problem to an equivalent (i.e. with the same
solution set) but smaller problem. Since complete inference is
too computationally expensive to be used in practice, inference methods
are often incomplete and enforce {\em local consistencies}.
A local consistency is a property of a CSP defined over ``local''
parts of the CSP, in other words defined over subsets of
the variables and constraints of the CSP. The main idea is to remove
from the domains of the variables the values that will not take part
of any solution. Such values are said to be {\em inconsistent}.
Inconsistent values can be detected by using a number of consistency
properties.





A common consistency property proposed in \cite{Mackworth77b} is
generalised arc-consistency. A constraint $c$ is \emph{generalised
arc-consistent} (or GAC), written GAC($c$), if and only if for every
$X\in$ {\em vars}($c$) and every $v\in\domain{D}(X)$, there is at
least one assignment to {\em vars}$(c)$ that assigns $v$ to $X$ and
satisfies $c$. Values for variables other than $X$ participating in
such assignments are known as the {\em support} for the assignment
of $v$ to $X$. Generalised arc-consistency is established on a
constraint $c$ by removing elements from the domains of variables in
{\em vars}($c$) until the GAC property holds. For binary
constraints, GAC is equivalent to {\em arc-consistency} (AC, see
\cite{Mackworth77a}). Another useful local consistency is
\emph{bound consistency}  that treats the domains of the variables
as intervals. For integer variables, the
values have a natural total order, therefore the domain can be
represented by an interval whose lower bound is the minimum value
and the upper bound is the maximum value in the domain. A constraint
$C$ is \emph{bound consistent} (\emph{BC}) iff for every variable,
for its minimum (maximum) there exists a value for every other
variable between its minimum and maximum that satisfies $C$
\cite{vanhentenryck2}.

We will compare local consistency properties applied to (sets of)
logically equivalent constraints, $c_1$ and $c_2$. As in
\cite{debruyne1}, we say that a local consistency property $\Phi$ on
$c_1$ is as strong as $\Psi$ on $c_2$ iff, given any domains, if
$\Phi$ holds on $c_1$ then $\Psi$ holds on $c_2$; we say that $\Phi$
on $c_1$ is strictly stronger than $\Psi$ on $c_2$
iff 
$\Phi$ on $c_1$ is as strong as $\Psi$ on $c_2$ but not vice versa.

In a constraint program, searching for solutions is interleaved with
local consistency as follows. Local consistency is first enforced
before search starts to preprocess the problem and prune subsequent
search. It is then maintained dynamically at each node of the search
tree with respect to the current variable assignment. In this way,
the domains of the uninstantiated variables shrink and the search
tree gets smaller.
%
Whilst the process of maintaining local consistency over a CSP is
known as {\em propagation}, the process of removing inconsistent
values from the domains is known as {\em pruning} or {\em
filtering}. For effective constraint solving, it is important that
propagation removes efficiently as many inconsistent values as
possible. Note that GAC is an important consistency property as it
is the strongest filtering that be done by reasoning on only a
single constraint at a time. Many global constraints in CP toolkits
therefore encapsulate their own {\em filtering algorithm} which
typically achieves GAC at a low cost by exploting the semantics of
the constraint. As an example,  R\'{e}gin in \cite{regin1} gives a
filtering algorithm for the $\myalldiff$ constraint which maintains
GAC in time $O(n^{2.5})$ where $n$ is the number of variables.

The semantics of a constraint can help not only find supports
and inconsistent values quickly but also detect entailment and
disentailment without having to do filtering. A constraint $c$ is
{\em entailed} if all assignments of values to {\em vars}$(c)$
satisfy $c$. Similarly, a constraint $c$ is {\em disentailed} when
all assignments of values to {\em vars}$(c)$ violate $c$. If a
constraint in a CSP is detected to be entailed, it does not have to
propagated in the future, and if it is detected to be disentailed
then it is proven that the current CSP has no solution and we can backtrack.



\subsection{Modelling}

CP toolkits provide constructs for declaring the variables, their
domains, as well as the constraints between these variables of a
CSP. They often contain a library of predefined constraints with a
particular semantics that can be applied to sets of variables with
varying arities and domains.
For instance, \myalldiff$([X_1,..,X_3])$ with
$D(X_1)=D(X_2)=\{1,2\},D(X_3)=\{1,2,3\}$ is an instance of
\myalldiff$([X_1,\ldots,X_n])$ defined on three variables with the
specified domains. It has the semantics that the variables involved
take different values \cite{regin1}. The
\myalldiff$([X_1,\ldots,X_n])$ constraint can be applied to any
number of variables with any domains. Such constraints are often
referred as {\em global constraints}. Beldiceanu has catalogued
hundreds of global constraints, most of which are defined over finite
domain variables \cite{Beldiceanu:catalog}. They permit the user to
model a problem easily by compactly specifying common patterns that
occur in many constraint models. They also provide solving
advantages which we shall explain later.



Since constraints provide a rich language, a number of alternative
models will often exist, some of which will be more effective than
others. However, one of the most common and effective modelling
patterns in constraint programming is a {\em matrix model}. A matrix
model is the formulation of a CSP with one or more matrices of
decision variables (of one or more dimensions) \cite{Reform02}.
Matrix models are a natural way to represent problems that involve
finding a function or a relation. We shall illustrate
matrix models and the power of global constraints in modelling
through the {\bf sport scheduling problem}.
\begin{figure}[t]
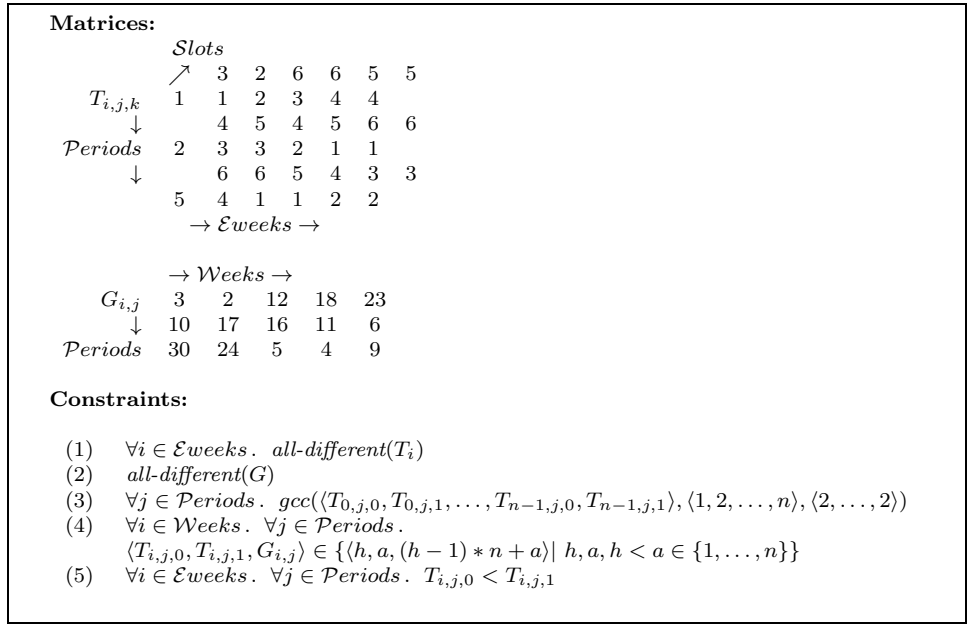

\begin{center}
\framebox{
\begin{scriptsize}
\begin{tabular}{l}
\mbox{{\bf Matrices:}} \\
$
\begin{array}{rcccccccc}
& \multicolumn{2}{c}{\mbox{\mydomain{Slots}}}\\
&\nearrow& 3 & 2 & 6 & 6 & 5 & 5 \\
T_{i,j,k}  &  1 & 1 & 2 & 3 & 4 & 4 \\
\downarrow&& 4 & 5 & 4 & 5 & 6 & 6\\
\mydomain{Periods} &  2 & 3 & 3 & 2 & 1 & 1 \\
\downarrow      && 6 & 6 & 5 & 4 & 3 & 3\\
                &  5 & 4 & 1 & 1 & 2 & 2 \\
& \multicolumn{5}{c}{\mbox{$\rightarrow \mydomain{Eweeks} \rightarrow$}}  \\
\end{array}$\\ \\
$
\begin{array}{rccccccc}
& \multicolumn{6}{c}{\mbox{$\rightarrow \mydomain{Weeks} \rightarrow$ }} \\
G_{i,j}      &3&2&12&18&23\\
\downarrow   &10&17&16&11&6\\
\mydomain{Periods}&30&24&5&4&9\\
\end{array}$\\

\\
\mbox{\bf Constraints:} \\\\
\begin{tabular}{ll}
(1) & $\forall i \in \mydomain{Eweeks} \Dot~ \myalldiff(T_{i})$ \\
(2) & $\myalldiff(G)$ \\
(3) & $\forall j \in \mydomain{Periods} \Dot~ gcc(\la T_{0,j,0}, T_{0,j,1},\ldots,T_{n-1,j,0}, T_{n-1,j,1}\ra, \la 1, 2, \ldots, n \ra, \la 2, \ldots, 2\ra)$\\
(4) & 
$\forall i \in \mydomain{Weeks} \Dot~ \forall j \in \mydomain{Periods} \Dot~$ \\
& $\la T_{i,j,0},T_{i,j,1},G_{i,j} \ra \in \{\la h,a,(h-1)*n+a \ra|~h,a,h<a \in \{1,\ldots,n\}\}$ \\
(5) & $\forall i \in \mydomain{Eweeks} \Dot~ \forall j \in
\mydomain{Periods} \Dot~ T_{i,j,0}<T_{i,j,1}$
\end{tabular} \\
\\
\end{tabular}
\end{scriptsize}
}

\end{center}
\caption{The matrix model of the sport scheduling problem in
\cite{vanhentenryck1}.} \label{ch-matrixmodels:table-sport}

\end{figure}
This problem involves scheduling games between $n$ teams over $n-1$
weeks \cite{vanhentenryck1}. Each week is divided into $n/2$
periods, and each period is divided into two slots. The team in the
first slot plays at home, while the team in the second slot plays
away. The goal is to find a schedule such that: (i) every team plays
exactly once a week; (ii) every team plays against every other team;
(iii) every team plays at most twice in the same period over the
tournament. Van Hentenryck {\em et al.} propose a model for this
problem in \cite{vanhentenryck1}, where they extend the problem with
a ``dummy'' final week to make the problem more uniform. The model
consists of two matrices: a 3-d matrix $T$ of $\mydomain{Periods}
\times \mydomain{Eweeks} \times \mydomain{Slots}$ and a 2-d matrix
$G$ of $\mydomain{Periods} \times \mydomain{Weeks}$, where
$\mydomain{Periods}$ is the set of $n/2$ periods,
$\mydomain{Eweeks}$ is the set of $n$ extended weeks,
$\mydomain{Weeks}$ is the set of $n-1$ weeks, and $\mydomain{Slots}$
is the set of $2$ slots. In $T$, weeks are extended to include the
dummy week, and each element takes a value from $\{1, \ldots, n\}$
expressing that a team plays in a particular week in a particular
period, in the home or away slot. For the sake of simplicity, we
will treat this matrix as 2-d where the rows represent the periods
and the columns represent the extended weeks, and each entry of the
matrix is a pair of variables. The elements of $G$ takes values from
$\{1,\ldots,n^2\}$, and each element denotes a particular unique
combination of home and away teams. More precisely, a game played
between a home team $h$ and an away team $a$ is uniquely identified
by $(h-1)*n+a$. (see Figure \ref{ch-matrixmodels:table-sport}).

Consider the columns of $T$ which denote the (extended) weeks. The
first set of constraints post   $\myalldiff$ (global) constraints
on the columns  of $T$ to enforce that each column is a permutation
of $1 \ldots n$.
The second constraint is an $\myalldiff$ (global) constraint on $G$
that enforces that  all games must be different.
Consider the rows of $T$ which represent the periods. The third set
of constraints post the global cardinality constraints ({\em gcc})
on the rows to ensure that each of $1 \ldots n$ occur exactly twice
in every row.
The fourth set of constraints are called {\em channelling
constraints} and are often used when multiple matrices are used to
model the problem and they have to be  linked together. In our case,
the channelling constraints links a variable representing a game
($G_{i,j}$) with a variable representing the team playing home team
($T_{i,j,0}$) and the corresponding variable representing the away
team ($T_{i,j,1}$) such that $G_{i,j} = (T_{i,j,0}-1)*n+T_{i,j,1}$.
The final set of constraints will be discussed after giving an
overview of symmetry in CP.

\subsection{Symmetry}

A {\em symmetry} is an intrinsic property of an object which is
preserved under certain classes of transformations. For instance,
rotating a chess board $90^\circ$ gives us a board which is
indistinguishable from the original one. A CSP can
have symmetries in the variables or domains or
both which preserve satisfiability. In the presence of symmetry, any
(partial) assignment can be transformed into a set of symmetrically
equivalent assignments without affecting whether or not the original
assignment satisfies the constraints.


Symmetry in constraint programs increases the size of
the search space. It is
therefore important to prune symmetric states so as to improve
the search efficiency. This process is referred to as {\em symmetry
breaking}. One of the easiest and most efficient ways of symmetry
breaking is adding \emph{symmetry breaking constraints}
\cite{puget:ismis93,clgrkr96}. These constraints impose an ordering
on the symmetric objects. Among the set of symmetric assignments,
only those that satisfy the ordering constraints are chosen for
consideration during the process of search. For instance, in the
matrix model of Figure~\ref{ch-matrixmodels:table-sport}, any
solution can be mapped to a symmetric solution by swapping any two
teams ($T_{i,j,0}$ and $T_{i,j,1}$). These solutions are essentially
the same. We can add the set of constraints (5) in
order to break such symmetry between the two teams and speed up
search by avoiding visiting symmetric branches.


A common pattern of symmetry in matrix models is that the rows
and/or columns of a 2-d matrix represent indistinguishable objects.
Consequently the rows and/or columns of an assignment
can be swapped without
affecting whether or not the assignment is a solution
\cite{ffhkmpwcp2002}. These are called \emph{row} or \emph{column
symmetry}; the general term is \emph{index symmetry}. For instance,
in the matrix model of Figure~\ref{ch-matrixmodels:table-sport}, the
(extended) weeks over which the tournament is held, as well the
periods of a week are indistinguishable. The rows and the columns of
$T$ and $G$ are therefore symmetric. Note that we treat $T$ as a 2-d
matrix where the rows represent the periods and columns represent
the (extended) weeks, and each entry of the matrix is a pair of
variables.

If every bijection on the values of an index is an index symmetry,
then we say that the index has \emph{total symmetry}. If the first
(resp. second) index of a 2-d matrix has total symmetry, we say that
the matrix has \emph{total column symmetry} (resp. \emph{total row
symmetry}). In many matrix models only a subset of the rows or
columns are interchangeable. If the first (resp. second) index of a
2-d matrix has partial symmetry, we say that the matrix has
\emph{partial column symmetry} (resp. \emph{partial row
symmetry})\footnote{Throughout, we will say {\em row symmetry}
(resp. \emph{column symmetry}) when we mean {\em total row
symmetry} (resp. \emph{total column symmetry}) to the problem
variables.}. There is one final case to consider: an index may have
partial index symmetry on multiple subsets of its values. For
example, a CSP may have a 2-d matrix for which rows 1, 2 and 3 are
interchangeable and rows 5, 6 and 7 are interchangeable. This can
occur on any or all of the indices.

An $n \times m$ matrix with total row and column symmetry has $n!m!$
symmetries, a number which increases super-exponentially. An effective way to
deal with this class of symmetry is to use lexicographic ordering
constraints.

\begin{definition}
\label{lexdef1} A strict lexicographic ordering  $\myvec{x} <_{lex}
\myvec{y}$ between two vectors of integers $\myvec{x}=\la x_0, x_1,
\ldots, x_{n-1} \ra$ and $\myvec{y}=\la y_0, y_1, \ldots, y_{n-1}
\ra$ holds iff $\exists k ~0 \leq k <n$ such that $x_i=y_i$ for all
$0 \leq i < k$ and $x_k<y_k$.
\end{definition}
The ordering can be weakened to include equality.
\begin{definition}
\label{lexdef2} Two vectors of integers $\myvec{x}=\la x_0, x_1,
\ldots, x_{n-1} \ra$ and $\myvec{y}=\la y_0, y_1, \ldots, y_{n-1}
\ra$ are lexicographically ordered $\myvec{x} \leq_{lex} \myvec{y}$
iff $\myvec{x} <_{lex} \myvec{y}$ or $\myvec{x}=\myvec{y}$.
 \end{definition}
Given two vectors of variables $\myvec{X}=\la X_0, X_1, \ldots,
X_{n-1} \ra$ and $\myvec{Y}=\la Y_0, Y_1, \ldots, Y_{n-1} \ra$, we
write a lexicographic ordering constraint as $\myvec{X} \leq_{lex}
\myvec{Y}$ and a strict lexicographic ordering constraint as
$\myvec{X} <_{lex} \myvec{Y}$. These constraints are satisfied by an
assignment if the vectors $\myvec{x}$ and $\myvec{y}$ assigned to
$\myvec{X}$ and $\myvec{Y}$ are ordered according to
Definitions~\ref{lexdef2} and \ref{lexdef1}, respectively.

To deal with column (resp. row) symmetry, we can constrain the
columns (resp. rows) to be non-decreasing as the value of the index
increases. One way to achieve this is by imposing a lexicographic
ordering constraint between adjacent columns (resp. rows). These
constraints are {\em consistent} which means that they leave at
least one assignment among the set of symmetric assignments. We can
deal with row and column symmetry in a similar way by imposing a
lexicographic ordering constraint between adjacent rows and columns
simultaneously. Also such constraints are consistent. Even though
these constraints may not eliminate all symmetry, they have been
shown to be effective at removing many symmetries from the search
spaces of many problems. If a matrix has only partial column (resp.
partial row) symmetry then the symmetry can be broken by
constraining the interchangeable columns (resp. rows) to be in
lexicographically non-decreasing order.  This can be achieved in a
manner similar to that described above.  The method also extends to
matrices that have partial or total column symmetry together with
partial or total row symmetry. Finally, if the columns and/or rows
of a matrix have multiple partial symmetries than each can be broken
in the manner just described \cite{ffhkmpwcp2002}.

%
%

\section{The Multiset Ordering Constraint and Its Applications}
\label{sec:App}

Multiset ordering is a total ordering on multisets.

\begin{definition}
\label{msetdef1} Strict multiset ordering \index{Strict multiset
ordering|see {Multiset ordering}} \index{Multiset ordering!strict}
$\mset{x} <_{m} \mset{y}$ between two multisets of integers
$\mset{x}$ and $\mset{y}$ holds iff:
\begin{eqnarray*}
 \mset{x}=\msetl \msetr~\And~\mset{y} \neq \msetl \msetr &  \Or & \\
 max(\mset{x})<max(\mset{y}) & \Or & \\
 (max(\mset{x})=max(\mset{y})~\And~\mset{x}-\msetl max(\mset{x}) \msetr <_m \mset{y}-\msetl max(\mset{y})\msetr) &&
\end{eqnarray*}
\end{definition}
That is, either $\mset{x}$ is empty and $\mset{y}$ is not, or the
largest value in $\mset{x}$ is less than the largest value in
$\mset{y}$, or the largest values are the same and, if we
eliminate one occurrence of the largest value from both $\mset{x}$
and $\mset{y}$, the resulting two multisets are ordered. We can
weaken the ordering to include multiset equality.
\begin{definition}
\label{msetdef2} Two multisets of integers $\mset{x}$ and
$\mset{y}$ are multiset ordered \index{Multiset ordering}
$\mset{x} \leq_{m} \mset{y}$ iff $\mset{x} <_{m} \mset{y}$ or
$\mset{x}=\mset{y}$.
 \end{definition}

Even though this ordering is defined on multisets, it may also be
useful to order vectors by ignoring the positions but rather
concentrating on the values taken by the variables. We can do this
by treating a vector as a multiset. Given two vectors of variables
$\myvec{X}=\la X_0, X_1, \ldots, X_{n-1} \ra$ and $\myvec{Y}=\la
Y_0, Y_1, \ldots, Y_{n-1} \ra$, we write a multiset ordering
constraint as $\myvec{X} \leq_{m} \myvec{Y}$ and a strict multiset
ordering constraint as $\myvec{X} <_{m} \myvec{Y}$. These
constraints ensure that the vectors $\myvec{x}$ and $\myvec{y}$
assigned to $\myvec{X}$ and $\myvec{Y}$, when viewed as multisets,
are multiset ordered according to Definitions \ref{msetdef2} and
\ref{msetdef1}, respectively.

\subsection{Breaking Index Symmetry}


One important application of the multiset ordering constraint is in
breaking index symmetry
\cite{fhkmw:ijcai03}. If $X$ is an $n$ by $m$ matrix of
decision variables, then we can break its column symmetry by
imposing the constraints $ \la X_{i,0}, \ldots, X_{i,m} \ra \leq_{m}
\la X_{i+1,0}, \ldots, X_{i+1,m} \ra$ for $i \in [0,n-2]$, or for
short $\myvec{C_0} \leq_{m} \myvec{C_{1}} \ldots \leq_{m} \myvec
C_{n-1} $ where $\myvec{C_i}$ corresponds to the vector of variables
$\la X_{i,0}, \ldots, X_{i,m} \ra$ which belong to the $i^{th}$
column of the matrix. Similarly we can break its row symmetry by
imposing the constraints $ \la X_{0,j}, \ldots, X_{n,j} \ra \leq_{m}
\la X_{0,j+1}, \ldots, X_{n,j+1} \ra$ for $j \in [0, m-2]$, or for
short $\myvec{R_0} \leq_{m} \myvec{R_{1}} \ldots \leq_{m} \myvec
R_{m-1} $ in which $\myvec{R_j}$ corresponds to the variables $\la
X_{0,j}, \ldots, X_{n,j} \ra$ of the $j^{th}$ row. Such constraints
are consistent symmetry breaking constraints. Note that when we have
partial column (resp. row) symmetry, then the symmetry can be broken
by  imposing multiset ordering constraints on the symmetric columns
(resp. rows) only.

Whilst multiset ordering is a total ordering on multisets, it is not
a total ordering on vectors. In fact, it is a preordering as it is
not antisymmetric. Hence, each symmetry class may have more than one
element where the rows (resp. columns) are multiset ordered. This
does not however make lexicographic ordering constraints preferable
over multiset ordering constraints in breaking row (resp. column)
symmetry. The reason is that
they are incomparable as they remove different symmetric assignments
in an equivalence class \cite{fhkmw:ijcai03}.

One of the nice features of using multiset ordering for breaking
index symmetry is that by constraining one dimension of the matrix,
say the rows, to be multiset ordered, we do not distinguish the
columns. We can still freely permute the columns, as multiset
ordering the rows ignores positions and is invariant to column
permutation. We can therefore consistently post multiset ordering
constraints on the rows together with either multiset ordering or
lexicographic ordering constraints on the columns when we have both
row and column symmetry. Neither approach may eliminate all
symmetries, however they are all potentially interesting. Since
lexicographic ordering and multiset ordering constraints are
incomparable, imposing one ordering in one dimension and the other
ordering in the other dimension of a matrix is also incomparable to
imposing the same ordering on both dimensions of the matrix
\cite{fhkmw:ijcai03}. Studying the effectiveness of all these
different methods in reducing index symmetry is outside the scope of
this paper as we only focus on the design of efficient and effective
filtering algorithms for the multiset ordering constraints.
Nevertheless, experimental results in Section
\ref{ch-gacmset:experiments} show that exploiting both multiset
ordering and lexicographic ordering constraints can be very
effective in breaking index symmetry.

A multiset ordering constraint can also be helpful for implementing
other constraints useful to break index symmetry. One such
constraint is {\em allperm} \cite{fjm:cp03}. Experimental results in
\cite{fjm:cp03} show that the decomposition of {\em allperm} using a
multiset ordering constraint can be as effective and efficient as
the specialised algorithm proposed.

\subsection{Searching for Leximin Optimal Solutions}


Another interesting application  of the multiset ordering constraint
arises in the context of searching for leximin optimal solutions.
Such solutions can be useful in fuzzy CSPs. A fuzzy constraint
associates a degree of satisfaction to an assignment tuple for the
variables it constrains. To combine degrees of satisfaction, we can
use a combination operator like the minimum function. Unfortunately,
the minimum function may cause a {\em drowning effect} when one
poorly satisfied constraint `drowns' many highly satisfied
constraints. One solution is to collect a vector of degrees of
satisfaction, sort these values in ascending order and compare them
lexicographically. This {\em leximin} combination operator
identifies the assignment that violates the fewest constraints
\cite{fargier:Thesis}.  This induces an ordering identical to the
multiset ordering except that the lower elements of the satisfaction
scale are the more significant.  It is simple to modify a multiset
ordering constraint to consider the values in a reverse order.  To
solve such leximin fuzzy CSPs, we can then use branch and bound,
adding a multiset ordering constraint when we find a solution to
ensure that future solutions are greater in the leximin ordering.

Leximin optimal solutions can be useful also in other domains. For
instance, as shown in \cite{leximin:ijcai07}, they can be exploited
as a fairness and pareto optimality criterion when solving
multiobjective problems in CP. Experimental results in
\cite{leximin:ijcai07} show that using a multiset ordering
constraint in a branch and bound search can be competitive with the
alternative approaches to finding leximin optimal solutions.

\section{A Filtering Algorithm for Multiset Ordering Constraint}
\label{ch-gacmset:algorithm}

In this section, we present our first filtering algorithm which
either detects that $\myvec{X} \leq_{m} \myvec{Y}$ is disentailed or
prunes inconsistent values so as to achieve GAC on $\myvec{X}
\leq_{m} \myvec{Y}$. After sketching the main features of the
algorithm on a running example in Section
\ref{ch-gacmset:algorithm-ex}, we first present the theoretical
results that the algorithm exploits in Section
\ref{ch-gacmset:algorithm-back} and then give the details of the
algorithm in Section \ref{ch-gacmset:algorithm-details}. Throughout,
we assume that the variables of the vectors  $\myvec{X}$ and
$\myvec{Y}$ are disjoint.

\subsection{A Worked Example}
\label{ch-gacmset:algorithm-ex} The key idea behind the algorithm
is to build a pair of occurrence vectors associated
with \myfloor($\myvec{X}$) and \myceiling($\myvec{Y}$).
The algorithm goes through every variable of $\myvec{X}$ and
$\myvec{Y}$ checking for support for values in the domains.
It suffices to have $occ(\myfloor(\myvec X_{X_i \assigned
max(X_i)})) \leq_{lex} occ(\myceiling(\myvec Y))$ to ensure that
all values of $\domain{D}(X_i)$ are consistent. Similarly, we only
need $occ(\myfloor(\myvec{X})) \leq_{lex} occ(\myceiling(\myvec
Y_{Y_j \assigned min(Y_j)}))$ to hold for the values of
$\domain{D}(Y_j)$ to be consistent. We can avoid the repeated
construction and traversal of these vectors by building, once and
for all, the vectors $occ(\myfloor(\myvec{X}))$ and
$occ(\myceiling(\myvec{Y}))$, and defining some pointers and flags
on them. For instance, assume we have $occ(\myfloor(\myvec{X}))
\leq_{lex} occ(\myceiling(\myvec{Y}))$. The vector
$occ(\myfloor(\myvec X_{X_i \assigned max(X_i)}))$ can be obtained
from $occ(\myfloor(\myvec{X}))$ by decreasing the number of
occurrences of $min(X_i)$ by 1, and increasing the number of
occurrences of $max(X_i)$ by 1. The pointers and flags tell us
whether this disturbs the lexicographic ordering, and if so they
help us to find quickly the largest $max(X_i)$ which does not.

Consider the multiset ordering constraint $\myvec X \leq_{m}
\myvec Y$ where:
\[
\begin{array}{ccccccccc}
 \myvec{X} &=& \la \{5\}, &\{4,5\}, &\{3,4,5\}, &\{2,4\}, &\{1\}, &\{1\}\ra\\
 \myvec{Y} &=& \la \{4,5\}, &\{4\}, &\{1,2,3,4\}, &\{2,3\}, &\{1\},&\{0\}\ra
\end{array}
\]
We have $\myfloor(\myvec X)=\la 5, 4, 3, 2, 1, 1\ra$ and
$\myceiling(\myvec Y)=\la 5, 4, 4, 3, 1, 0 \ra$. We construct our
occurrence vectors $\myvec{ox}=occ(\myfloor(\myvec X))$ and
$\myvec{oy}=occ(\myceiling(\myvec Y))$, indexed from $max( \msetl
\myceiling(\myvec{X}) \msetr \cup \msetl \myceiling(\myvec{Y})
\msetr)=5$ to $min( \msetl \myfloor(\myvec{X}) \msetr \cup \msetl
\myfloor(\myvec{Y}) \msetr)=0$:
\[
\begin{array}{cccccccc}
  &   &     5   &4  &3    &2   &1   &0\\
 \myvec{ox} & = & \la 1, & 1, &  1, & 1, & 2, & 0 \ra \\
 \myvec{oy} & = & \la 1, & 2, &  1, & 0, & 1, & 1 \ra
\end{array}
\]
Recall that ${ox}_i$ and ${oy}_i$ denote the number of occurrences
of the value $i$ in $\msetl \myfloor(\myvec{X}) \msetr$ and
$\msetl \myceiling(\myvec{Y}) \msetr$, respectively. For example,
$oy_4=2$ as $4$ occurs twice in $\msetl \myceiling(\myvec{Y})
\msetr$. Next, we define our pointers and flags on $\myvec{ox}$
and $\myvec{oy}$. The pointer $\alpha$ points to the most
significant index above which the values are pairwise equal and at
$\alpha$ we have $ox_\alpha<oy_\alpha$. This means that we will
fail to find support if any of the $X_i$ is assigned a new value
greater than $\alpha$, but we will always find support for values
less than $\alpha$. If $\myvec{ox}=\myvec{oy}$ then we set
$\alpha=-\infty$. Otherwise, we fail immediately because no value
for any variable can have support. We define $\beta$ as the most
significant index below $\alpha$ such that $ox_\beta> oy_\beta$.
This means that we might fail to find support if any of the $Y_j$
is assigned a new value less than or equal to $\beta$, but we will
always find support for values larger than $\beta$. If such an
index does not exist then we set $\beta=-\infty$. Finally, the
flag $\gamma$ is $true$ iff $\beta=\alpha-1$ or
$\myvec{ox}_{\alpha+1 \rightarrow \beta-1}=\myvec{oy}_{\alpha+1
\rightarrow \beta-1}$, and $\sigma$ is $true$ iff the subvectors
below $\beta$ are ordered lexicographically the wrong way. In our
example, $\alpha=4$, $\beta=2$, $\gamma=true$, and $\sigma=true$:
\[
\begin{array}{cccccccc}
  &   &     5   &4  &3    &2   &1   &0\\
 \myvec{ox} & = & \la 1, & 1, &  1, & 1, & 2, & 0 \ra \\
 \myvec{oy} & = & \la 1, & 2, &  1, & 0, & 1, & 1 \ra \\
            &   &        &\alpha \uparrow &\gamma =true & \beta \uparrow  & \sigma=true&
\end{array}
\]
We now go through each $X_i$ and find the largest value in its
domain which is supported. If $X_i$ has a singleton domain then we
skip it because we have $\myvec{ox} \leq_{lex} \myvec{oy}$,
meaning that its only value has support.
%
%
Consider $X_1$. As $min(X_1) = \alpha$, changing $\myvec{ox}$ to
$occ(\myfloor(\myvec X_{X_1 \assigned max(X_1)}))$ increases the
number of occurrences of an index above $\alpha$ by $1$. This upsets
$\myvec{ox} \leq_{lex} \myvec{oy}$. We therefore prune all values in
$\domain{D}(X_1)$ larger than $\alpha$. Now consider $X_2$. We have
$max(X_2) > \alpha$ and $min(X_2)<\alpha$. As with $X_1$, any value
of $X_2$ larger than $\alpha$ upsets the lexicographic ordering, but
any value less than $\alpha$ guarantees the lexicographic ordering.
The question is whether $\alpha$ has any support? Changing
$\myvec{ox}$ to $occ(\myfloor(\myvec X_{X_2 \assigned \alpha}))$
decreases the number of occurrences of $3$ in $\myvec{ox}$ by 1, and
increases the number of occurrences of $\alpha$ by $1$. Now we have
$ox_\alpha=oy_\alpha$ but decreasing an entry in $\myvec{ox}$
between $\alpha$ and $\beta$ guarantees lexicographic ordering. We
therefore prune from $\domain{D}(X_2)$ only the values greater than
$\alpha$. Now consider $X_3$. We have $max(X_3)=\alpha$ and
$min(X_3)<\alpha$. Any value less than $\alpha$ has support but does
$\alpha$ have any support? Changing $\myvec{ox}$ to
$occ(\myfloor(\myvec X_{X_3 \assigned \alpha}))$ decreases the
number of occurrences of $beta$ in $\myvec{ox}$ by 1, and increases
the number of occurrences of $\alpha$ by 1. Now we have
$ox_\alpha=oy_\alpha$ and $ox_\beta=oy_\beta$. Since $\gamma$ and
$\sigma$ are $true$, the occurrence vectors are lexicographically
ordered the wrong way. We therefore prune $\alpha$ from
$\domain{D}(X_3)$. We skip $X_4$ and $X_5$.

Similarly, we go through each $Y_j$ and find the smallest value in
its domain which is supported. If $Y_j$ has a singleton domain
then we skip it because we have $\myvec{ox} \leq_{lex}
\myvec{oy}$, meaning that its only value has support. Consider
$Y_0$. As $max(Y_0)
> \alpha$, changing $\myvec{oy}$ to $occ(\myceiling(\myvec
Y_{Y_0 \assigned min(Y_0)}))$  decreases the number of occurrences
of an index above $\alpha$ by $1$. This upsets $\myvec{ox}
\leq_{lex} \myvec{oy}$. We therefore prune all values in
$\domain{D}(Y_0)$ less than or equal to $\alpha$. Now consider
$Y_2$. We have $max(Y_2)=\alpha$ and $min(Y_2) \leq \beta$. Any
value larger than $\beta$ guarantees lexicographic ordering. The
question is whether the values less than or equal to $\beta$ have
any support? Changing $\myvec{oy}$ to $occ(\myceiling(\myvec Y_{Y_2
\assigned min(Y_2)}))$ decreases the number of occurrences of
$\alpha$ by 1, giving us $ox_\alpha=oy_\alpha$. If $min(Y_2)=\beta$
then we have $ox_\beta=oy_\beta$. This disturbs $\myvec{ox}
\leq_{lex} \myvec{oy}$ because $\gamma$ and $\sigma$ are both
$true$. If $min(Y_2)<\beta$ then again we disturb $\myvec{ox}
\leq_{lex} \myvec{oy}$ because $\gamma$ is $true$ and the vectors
are not lexicographically ordered as of $\beta$. So, we prune from
$\domain{D}(Y_2)$ the values less than or equal to $\beta$. Now
consider $Y_3$. As $max(Y_3) < \alpha$, changing $\myvec{oy}$ to
$occ(\myceiling(\myvec Y_{Y_3 \assigned min(Y_3)}))$ does not change
that $\myvec{ox} \leq_{lex} \myvec{oy}$. Hence, $min(Y_3)$ is
supported. We skip $Y_4$ and $Y_5$.

We have now the following generalised arc-consistent vectors:
\[
\begin{array}{cccccccc}
 \myvec{X} &=& \la \{5\}, &\{4\}, &\{3,4\}, &\{2\}, &\{1\}, &\{1\} \ra\\
 \myvec{Y} &=& \la \{5\}, &\{4\}, &\{3,4\}, &\{2,3\}, &\{1\}, & \{0\}\ra
\end{array}
\]

\subsection{Theoretical Background}
\label{ch-gacmset:algorithm-back}

The algorithm exploits four theoretical results. The first reduces
GAC to consistency on the upper bounds of $\myvec{X}$ and on the
lower bounds of $\myvec{Y}$. The second and the third show in turn
when $\myvec{X} \leq_m \myvec{Y}$ is disentailed and what conditions
ensure GAC on $\myvec{X} \leq_m \myvec{Y}$. And the fourth
establishes that two ground vectors are multiset ordered iff the
associated occurrence vectors are lexicographically ordered.

\begin{theorem}
\label{th-BCmset} GAC($\myvec X \leq_m \myvec
Y$)\index{Arc-consistency!generalised} iff for all $0 \leq i < n$,
$max(X_i)$ and $min(Y_i)$ are consistent.
\end{theorem}
\proof GAC implies that every value is consistent. To show the
reverse, suppose  for all $0 \leq i < n$, $max(X_i)$ and
$min(Y_i)$ are supported, but the constraint is not GAC. Then
there is an inconsistent value. If this value is in some
$\domain{D}(X_i)$ then any value greater than this value, in
particular $max(X_i)$, is inconsistent. Similarly, if the
inconsistent value is in some $\domain{D}(Y_i)$ then any value
less than this value, in particular $min(Y_i)$, is inconsistent.
In any case, the bounds are not consistent. \qed

A constraint is said to be {\em disentailed} when the constraint is
$false$. The next two theorems show when $\myvec X \leq_m \myvec Y$
is disentailed and what conditions ensure  GAC on $\myvec X \leq_m
\myvec Y$.

\begin{theorem}
\label{th-msetGAC1} $\myvec{X} \leq_m \myvec{Y}$ is
disentailed\index{Disentailment} iff $\msetl \myfloor(\myvec{X})
\msetr
>_m \msetl \myceiling(\myvec{Y} )\msetr$.
\end{theorem}
\proof ($\Rightarrow$) Since  $\myvec{X} \leq_m \myvec{Y}$ is
disentailed, any combination of assignments, including $\myvec{X}
\assigned \myfloor(\myvec{X})$ and $\myvec{Y} \assigned
\myceiling(\myvec{Y})$, does not satisfy $\myvec{X} \leq_{m}
\myvec{Y}$. Hence, $\msetl \myfloor(\myvec{X}) \msetr >_m \msetl
\myceiling(\myvec{Y} )\msetr$.

($\Leftarrow$) Any $\myvec{x}\in \myvec{X}$ is greater than any
$\myvec{y}\in \myvec{Y}$ under the multiset ordering. Hence,
$\myvec{X} \leq_m \myvec{Y}$ is disentailed. \qed

\begin{theorem}
\label{th-msetGAC2} GAC($\myvec{X} \leq_m
\myvec{Y}$)\index{Arc-consistency!generalised} iff for all $i$ in
$[0,n)$:
\begin{eqnarray}
 \label{th-msetGAC2-first} \msetl \myfloor(\myvec{X}_{X_i \assigned max(X_i)}) \msetr &\leq_{m}& \msetl \myceiling(\myvec{Y}) \msetr \\
 \label{th-msetGAC2-second}\msetl \myfloor(\myvec{X}) \msetr & \leq_{m}& \msetl \myceiling(\myvec{Y}_{Y_i \assigned min(Y_i)})\msetr
\end{eqnarray}
\end{theorem}
\proof ($\Rightarrow$) As the constraint is GAC, all values have
support. In particular, $X_i \assigned max(X_i)$ has a support
$\myvec{x_1} \in \XvecXmax$ and $\myvec{y_1} \in \myvec{Y}$ where
$\msetl \myvec{x_1} \msetr \leq_m \msetl \myvec{y_1} \msetr$. Any
$\myvec{x_2} \in \XvecXmax$ less than or equal to $\myvec{x_1}$,
and any $\myvec{y_2} \in \myvec{Y}$ greater than or equal to
$\myvec{y_1}$, under multiset ordering, support $X_i \assigned
max(X_i)$. In particular, $\minXvecXmax$ and $max\{\myvec{y} \mid
~\myvec{y}\in\myvec{Y}\}$ support $X_i \assigned max(X_i)$. We get
$\minXvecXmax$ if all the other variables in $\myvec{X}$ take
their minimums, and we get $max\{\myvec{y} \mid
~\myvec{y}\in\myvec{Y}\}$ if all the variables in $\myvec{Y}$ take
their maximums. Hence, $\msetl \myfloor(\myvec{X}_{X_i \assigned
max(X_i)}) \msetr \leq_{m} \msetl \myceiling(\myvec{Y}) \msetr$.

A dual argument holds for the variables of $\myvec{Y}$. As the
constraint is GAC, $Y_i \assigned min(Y_i)$ has a support
$\myvec{x_1} \in \myvec{X}$ and $\myvec{y_1} \in \YvecYmin$ where
$\msetl \myvec{x_1} \msetr \leq_m \msetl \myvec{y_1} \msetr$. Any
$\myvec{x_2} \in \myvec{X}$ less than or equal to $\myvec{x_1}$,
and any $\myvec{y_2} \in \YvecYmin$ greater than or equal to
$\myvec{y_1}$, in particular $min\{\myvec{x} \mid
~\myvec{x}\in\myvec{X}\}$ and $\maxYvecYmin$ support $Y_i
\assigned min(Y_i)$. We get $min\{\myvec{x} \mid
~\myvec{x}\in\myvec{X}\}$ if all the variables in $\myvec{X}$ take
their minimums, and we get $\maxYvecYmin$ if all the other
variables in $\myvec{Y}$ take their maximums. Hence, $ \msetl
\myfloor(\myvec{X}) \msetr \leq_{m} \msetl
\myceiling(\myvec{Y}_{Y_i \assigned min(Y_i)})\msetr $.

($\Leftarrow$) Equation (\ref{th-msetGAC2-first}) ensures that for
all $0 \leq i < n$, $max(X_i)$ is supported, and Equation
(\ref{th-msetGAC2-second}) ensures that for all $0 \leq i < n$,
$min(Y_i)$ is supported. By Theorem \ref{th-BCmset}, the constraint
is GAC. \qed

In Theorems \ref{th-msetGAC1} and \ref{th-msetGAC2}, we need to
check whether two ground vectors are multiset ordered. The
following theorem shows that we can do this by lexicographically
comparing the occurrence vectors associated with these vectors.

\begin{theorem}
\label{th-occ}$\msetl \myvec{x} \msetr \leq_m \msetl \myvec{y}
\msetr$ iff $occ(\myvec{x}) \leq_{lex} occ(\myvec{y})$.
\end{theorem}
\proof ($\Rightarrow$) Suppose $\msetl \myvec{x} \msetr = \msetl
\myvec{y} \msetr$. Then the occurrence vectors associated with
$\myvec{x}$ and $\myvec{y}$ are the same. Suppose $\msetl
\myvec{x} \msetr <_m \msetl \myvec{y} \msetr$. If $max\msetl
\myvec{x} \msetr<max\msetl \myvec{y} \msetr$ then the leftmost
index of $\myvec{ox}=occ(\myvec{x})$ and
$\myvec{oy}=occ(\myvec{y})$ is $max\msetl \myvec{y} \msetr$, and
we have $ox_{max\msetl \myvec{y} \msetr}=0$ and $oy_{max\msetl
\myvec{y} \msetr}>0$. This gives $\myvec{ox} <_{lex} \myvec{oy}$.
If $max\msetl \myvec{x} \msetr=max\msetl \myvec{y} \msetr=a$ then
we eliminate one occurrence of $a$ from each multiset and compare
the resulting multisets.

($\Leftarrow$) Suppose $occ(\myvec{x})= occ(\myvec{y})$. Then
$\msetl \myvec{x} \msetr$ and $\msetl \myvec{y} \msetr$ contain
the same elements with equal occurrences. Suppose $occ(\myvec{x})
<_{lex} occ(\myvec{y})$. Then a value $a$ occurs more in $\msetl
\myvec{y} \msetr$ than in $\msetl \myvec{x} \msetr$, and the
occurrence of any value $b>a$ is the same in both multisets. By
deleting all the occurrences of $a$ from $\msetl \myvec{x} \msetr$
and the same number of occurrences of $a$  from $\msetl \myvec{y}
\msetr$, as well as any $b>a$ from both multisets, we get
$max\msetl \myvec{x} \msetr<max\msetl \myvec{y} \msetr$. \qed

Theorems \ref{th-msetGAC1} and \ref{th-msetGAC2} together with
Theorem \ref{th-occ} yield to the following propositions:

\begin{proposition}
\label{pr-msetGAC1}$\myvec{X} \leq_m \myvec{Y}$ is
disentailed\index{Disentailment} iff $occ( \myfloor(\myvec{X}))
>_{lex} occ(\myceiling(\myvec{Y}))$.
\end{proposition}

\begin{proposition}
\label{pr-msetGAC2} GAC($\myvec{X} \leq_m
\myvec{Y}$)\index{Arc-consistency!generalised} iff for all $i$ in
$[0,n)$:
\begin{eqnarray}
 \label{pr-msetGAC2-first}occ(\myfloor(\myvec{X}_{X_i \assigned max(X_i)})) &\leq_{lex}& occ( \myceiling(\myvec{Y}))\\
 \label{pr-msetGAC2-second}occ(\myfloor(\myvec{X})) & \leq_{lex}& occ(\myceiling(\myvec{Y}_{Y_i \assigned
 min(Y_i)}))
\end{eqnarray}
\end{proposition}


A naive way to enforce GAC on $\myvec{X} \leq_m \myvec{Y}$ is going
through every variable in the vectors, constructing the appropriate
occurrence vectors, and checking if their bounds satisfy
\ref{pr-msetGAC2-first} and \ref{pr-msetGAC2-second}. If they do,
then the bound is consistent. Otherwise, we try the nearest bound
until we obtain a consistent bound. We can, however, do better than
this by building only the vectors $occ(\myfloor(\myvec{X}))$ and
$occ(\myceiling(\myvec{Y}))$, and then defining some pointers and
Boolean flags on them. This saves us from the repeated construction
and traversal of the appropriate occurrence vectors. Another
advantage is that we can find consistent bounds without having to
explore the values in the domains.

We start by defining our pointers and flags. We write $\myvec{ox}$
for $occ(\myfloor(\myvec{X}))$, and $\myvec{oy}$ for
$occ(\myceiling(\myvec{Y}))$. We assume $\myvec{ox}$ and
$\myvec{oy}$ are indexed from $u$ to $l$, and $\myvec{ox} \leq_{lex}
\myvec{oy}$.\footnote{In the context of occurrence vector indexing,
$u..l$ and $[u,l]$ imply $u \geq l$. The exact meaning of the these
abused notations will be clear from the context.}

\begin{definition}
\label{ch:gacmset-def-alpha} Given
$\myvec{ox}=occ(\myfloor(\myvec{X}))$ and
$\myvec{oy}=occ(\myceiling(\myvec{Y}))$ indexed as $u..l$ where
$\myvec{ox} \leq_{lex} \myvec{oy}$, the pointer
$\alpha$\index{alpha@$\alpha$} is set either to the index in
$[u,l]$ such that:
\[
ox_\alpha < oy_\alpha~\And
\]
\[
\forall i~ u  \geq i > \alpha \Dot ox_i = oy_i
\]
or (if this is not the case) to $-\infty$.
\end{definition}
Informally, $\alpha$ points to the most significant index in
$[u,l]$ such that $ox_\alpha<oy_\alpha$ and all the variables
above it are pairwise equal. If, however, $\myvec{ox} =\myvec{oy}$
then $\alpha$ points to $-\infty$.

\begin{definition}
\label{ch:gacmset-def-beta} Given
$\myvec{ox}=occ(\myfloor(\myvec{X}))$ and
$\myvec{oy}=occ(\myceiling(\myvec{Y}))$ indexed as $u..l$ where
$\myvec{ox} \leq_{lex} \myvec{oy}$, the pointer
$\beta$\index{beta@$\beta$} is set either to the index in
$(\alpha,l]$ such that:
\[
ox_\beta > oy_\beta~\And
\]
\[
\forall i~ \alpha  > i > \beta \Dot ox_i \leq oy_i
\]
or (if $\alpha \leq l$ or for all $\alpha>i \geq l$ we have $ox_i
\leq oy_i$) to $-\infty$.
\end{definition}
Informally, $\beta$ points to the most significant index in
$(\alpha,l]$ such that $\myvec{ox}_{\beta \rightarrow l}
>_{lex} \myvec{oy}_{\beta \rightarrow l}$. If, such an index does not exist, then
$\beta$ points to $-\infty$. Note that we have $\sum_i ox_i=\sum_i
oy_i=n$, as $\myvec{ox}$ and $\myvec{oy}$ are both associated with
vectors of length $n$. Hence, $\alpha$ cannot be $l$, and we always
have $\myvec{ox}_{\alpha-1 \rightarrow l}
>_{lex} \myvec{oy}_{\alpha-1 \rightarrow l}$ when $\alpha \neq
-\infty$.

\begin{definition}
\label{ch:gacmset-def-gamma} Given
$\myvec{ox}=occ(\myfloor(\myvec{X}))$ and
$\myvec{oy}=occ(\myceiling(\myvec{Y}))$ indexed as $u..l$ where
$\myvec{ox} \leq_{lex} \myvec{oy}$, the flag
$\gamma$\index{gamma@$\gamma$} is $true$ iff:
\[
\beta \neq -\infty ~\And (\beta=\alpha-1~\Or~\forall i~ \alpha  >
i
> \beta \Dot ox_i = oy_i)
\]
\end{definition}
Informally, $\gamma$ is $true$ if $\beta \neq -\infty$, and either
$\beta$ is jut next to $\alpha$ or the subvectors between $\alpha$
and $\beta$ are equal. Otherwise, $\gamma$ is $false$.

\begin{definition}
\label{ch:gacmset-def-sigma} Given
$\myvec{ox}=occ(\myfloor(\myvec{X}))$ and
$\myvec{oy}=occ(\myceiling(\myvec{Y}))$ indexed as $u..l$ where
$\myvec{ox} \leq_{lex} \myvec{oy}$, the flag
$\sigma$\index{sigma@$\sigma$} is $true$ iff:
\[
\beta > l ~\And~\myvec{ox}_{\beta-1 \rightarrow l} >_{lex}
\myvec{oy}_{\beta-1 \rightarrow l}
\]
\end{definition}
Informally, $\sigma$ is $true$ if $\beta >l$ and the subvectors
below $\beta$ are lexicographically ordered the wrong way. If,
however, $\beta \leq l$, or the subvectors below $\beta$ are
lexicographically ordered, then $\sigma$ is $false$.

Using $\alpha$, $\beta$, $\gamma$, and $\sigma$, we can find the
tight upper bound for each $\domain{D}(X_i)$, as well as the tight
lower bound for each $\domain{D}(Y_i)$ without having to traverse
the occurrence vectors. In the next three theorems, we are
concerned with $X_i$. When looking for a support for a value $v
\in \domain{D}(X_i) $, we obtain $occ(\myfloor(\myvec X_{X_i
\assigned v}))$ by increasing $ox_v$ by 1, and decreasing
$ox_{min(X_i)}$ by 1. Since $\myvec{ox} \leq_{lex} \myvec{oy}$,
$min(X_i)$ is consistent. We therefore seek support for values
greater than $min(X_i)$.

\begin{theorem}
\label{th-X3} Given $\myvec{ox}=occ(\myfloor(\myvec{X}))$ and
$\myvec{oy}=occ(\myceiling(\myvec{Y}))$ indexed as $u..l$ where
$\myvec{ox} \leq_{lex} \myvec{oy}$, if $max(X_i) \geq \alpha$ and
$min(X_i)<\alpha$ then for all $v \in \domain{D}(X_i)$:
\begin{enumerate}
\item  if  $v >\alpha$ then $v$ is
inconsistent;
\item if $v <\alpha$ then $v$ is
consistent;
\item if $v=\alpha$ then $v$ is inconsistent iff:
\begin{eqnarray*}
 (ox_{\alpha}+1=oy_{\alpha}~\And~min(X_i)=\beta~\And~\gamma~\And~ox_{\beta}>oy_{\beta}+1) & ~\Or~ &\\
 (ox_{\alpha}+1=oy_{\alpha}~\And~min(X_i)=\beta~\And~\gamma~\And~ox_{\beta}=oy_{\beta}+1~\And~\sigma) & ~\Or~ &\\
 (ox_{\alpha}+1=oy_{\alpha}~\And~min(X_i)<\beta~\And~\gamma) &&
\end{eqnarray*}
\end{enumerate}
\end{theorem}
\proof If $min(X_i)<\alpha$ then $\alpha \neq -\infty$ and
$\myvec{ox}<_{lex} \myvec{oy}$. Let $v$ be a value in
$\domain{D}(X_i)$ greater than $\alpha$. Increasing $ox_{v}$ by 1
gives $\myvec{ox}
>_{lex} \myvec{oy}$. By Proposition \ref{pr-msetGAC2}, $v$ is inconsistent. Now let $v$
be less than $\alpha$. Increasing $ox_{v}$ by 1 does not change
$\myvec{ox} <_{lex} \myvec{oy}$. By Proposition \ref{pr-msetGAC2},
$v$ is consistent. Is $\alpha$ a tight upper bound? If any of the
conditions in item 3 is $true$ then we obtain $\myvec{ox} >_{lex}
\myvec{oy}$ by increasing $ox_{\alpha}$ by 1 and decreasing
$ox_{min(X_i)}$ by 1. By Proposition \ref{pr-msetGAC2}, $v=\alpha$
is inconsistent and therefore the largest value which is less than
$\alpha$ is the tight upper bound. We now need to show that the
conditions of item 3 are exhaustive. If $v=\alpha$ is inconsistent
then, by Proposition \ref{pr-msetGAC2}, we obtain $\myvec{ox}>_{lex}
\myvec{oy}$ after increasing $ox_{\alpha}$ by 1 and decreasing
$ox_{min(X_i)}$ by 1. This can happen only if
$ox_{\alpha}+1=oy_{\alpha}$ because otherwise we still have
$ox_{\alpha}<oy_{\alpha}$. 
Now, it is important where we decrease an occurrence. If it is above
$\beta$ (but below $\alpha$ as $min(X_i)<\alpha$) then we still have
$\myvec{ox} <_{lex} \myvec{oy}$ because for all
$\alpha>i>max\{l-1,\beta\}$, we have $ox_i \leq oy_i$. If it is on
or below $\beta$ (when $\beta \neq -\infty$) and $\gamma$ is
$false$, then we still have $\myvec{ox} <_{lex} \myvec{oy}$ because
$\gamma$ is $false$ when $\beta<\alpha-1$ and $\myvec{ox}_{\alpha-1
\rightarrow \beta+1} <_{lex} \myvec{ox}_{\alpha-1 \rightarrow
\beta+1}$. Therefore, it is necessary to have
$ox_{\alpha+1}+1=oy_{\alpha}~\And~min(X_i)\leq \beta~\And~\gamma$
for $\alpha$ to be inconsistent. Two cases arise here. In the first,
we have
$ox_{\alpha+1}+1=oy_{\alpha}~\And~min(X_i)=\beta~\And~\gamma$.
Decreasing $ox_{\beta}$ by 1 can give $\myvec{ox}>_{lex} \myvec{oy}$
in two ways: either we still have $ox_{\beta}>oy_{\beta}$, or we now
have $ox_{\beta}=oy_{\beta}$ but the vectors below $\beta$ are
ordered lexicographically the wrong way. Note that decreasing
$ox_{\beta}$ by 1 cannot give $ox_{\beta}<oy_{\beta}$. Therefore,
the first case results in two conditions for $\alpha$ to be
inconsistent:
$ox_{\alpha+1}+1=oy_{\alpha}~\And~min(X_i)=\beta~\And~\gamma~\And~ox_{\beta}>oy_{\beta}+1$
or
$ox_{\alpha+1}+1=oy_{\alpha}~\And~min(X_i)=\beta~\And~\gamma~\And~ox_{\beta}=oy_{\beta}+1~\And~\sigma$.
Now consider the second case, where we have
$ox_{\alpha+1}+1=oy_{\alpha}~\And~min(X_i)<\beta~\And~\gamma$.
Decreasing $ox_{min(X_i)}$ by 1 gives $\myvec{ox}>_{lex}
\myvec{oy}$. Hence, if $v=\alpha$ is inconsistent then we have
either of the three conditions. \qed

\begin{theorem}
\label{th-X1}Given $\myvec{ox}=occ(\myfloor(\myvec{X}))$ and
$\myvec{oy}=occ(\myceiling(\myvec{Y}))$ indexed as $u..l$ where
$\myvec{ox} \leq_{lex} \myvec{oy}$, if $max(X_i)<\alpha$ then
$max(X_i)$ is the tight upper bound.
\end{theorem}
\proof If $max(X_i)<\alpha$ then we have $\alpha \neq -\infty$ and
$\myvec{ox} <_{lex} \myvec{oy}$. Increasing $ox_{max(X_i)}$ by 1
does not change this. By Proposition \ref{pr-msetGAC2}, $max(X_i)$
is consistent. \qed

\begin{theorem}
\label{th-X2} Given $\myvec{ox}=occ(\myfloor(\myvec{X}))$ and
$\myvec{oy}=occ(\myceiling(\myvec{Y}))$ indexed as $u..l$ where
$\myvec{ox} \leq_{lex} \myvec{oy}$, if $min(X_i) \geq \alpha$ then
$min(X_i)$ is the tight upper bound.
\end{theorem}
\proof Any $v > min(X_i)$ in $\domain{D}(X_i)$ is greater than
$\alpha$. Increasing $ox_{v}$ by 1 gives $\myvec{ox}
>_{lex} \myvec{oy}$. By Proposition \ref{pr-msetGAC2}, any $v > min(X_i)$ in $\domain{D}(X_i)$ is inconsistent.
\qed

In the next four theorems, we are concerned with $Y_i$. When
looking for a support for a value $v \in \domain{D}(Y_i) $, we
obtain $occ(\myceiling(\myvec Y_{Y_i \assigned v}))$ by increasing
$oy_v$ by 1, and decreasing $oy_{max(Y_i)}$ by 1. Since
$\myvec{ox} \leq_{lex} \myvec{oy}$, $max(Y_i)$ is consistent. We
therefore seek support for values less than $max(Y_i)$.

\begin{theorem}
\label{th-Y4} Given $\myvec{ox}=occ(\myfloor(\myvec{X}))$ and
$\myvec{oy}=occ(\myceiling(\myvec{Y}))$ indexed as $u..l$ where
$\myvec{ox} \leq_{lex} \myvec{oy}$, if $max(Y_i) = \alpha$ and
$min(Y_i) \leq \beta$ then for all $v \in \domain{D}(Y_i)$
\begin{enumerate}
\item  if $v >\beta$ then $v$ is
consistent;
\item if $v <\beta$ then $v$ is
inconsistent iff $ox_{\alpha}+1=oy_{\alpha}~\And~\gamma$
\item if $v=\beta$ then $v$ is inconsistent iff:
\begin{eqnarray*}
 (ox_{\alpha}+1=oy_{\alpha}~\And~\gamma~\And~ox_{\beta}>oy_{\beta}+1) & ~\Or~ &\\
 (ox_{\alpha}+1=oy_{\alpha}~\And~\gamma~\And~ox_{\beta}=oy_{\beta}+1~\And~\sigma) & &
\end{eqnarray*}
\end{enumerate}
\end{theorem}
\proof If $max(Y_i)=\alpha$ and $min(Y_i)\leq \beta$ then $\alpha
\neq -\infty$, $\beta \neq -\infty$, and $\myvec{ox}<_{lex}
\myvec{oy}$. Let $v$ be a value in $\domain{D}(Y_i)$ greater than
$\beta$. Increasing $oy_{v}$ by 1 and decreasing $oy_{\alpha}$ by
1 does not change $\myvec{ox} <_{lex} \myvec{oy}$. This is because
for all $\alpha>i>\beta$, we have $ox_i \leq oy_i$. Even if now
$\myvec{ox}_{\alpha \rightarrow v+1}=\myvec{oy}_{\alpha
\rightarrow v+1}$, at $v$ we have $ox_v<oy_v$. By Proposition
\ref{pr-msetGAC2}, $v$ is consistent. Now let $v$ be less than
$\beta$. If the condition in item 2 is $true$ then we obtain
$\myvec{ox} >_{lex} \myvec{oy}$ by decreasing $oy_{\alpha}$ by 1
and increasing $oy_v$ by 1. By Proposition \ref{pr-msetGAC2}, $v$
is inconsistent. We now need to show that this condition is
exhaustive. If $v$ is inconsistent then by Proposition
\ref{pr-msetGAC2}, we obtain $\myvec{ox}>_{lex} \myvec{oy}$ after
decreasing $oy_{\alpha}$ by 1 and increasing $oy_{v}$ by 1. This
is in fact the same as obtaining $\myvec{ox}>_{lex} \myvec{oy}$
after increasing $ox_{\alpha}$ by 1 and decreasing $ox_{v}$ by 1.
We have already captured this case in the last condition of item 3
in Theorem \ref{th-X3}. Hence, it is necessary to have
$ox_{\alpha}+1=oy_{\alpha}~\And~\gamma$ for $v$ to be
inconsistent. What about $\beta$ then? If any of the conditions in
item 3 is $true$ then we obtain $\myvec{ox}
>_{lex} \myvec{oy}$ by decreasing $oy_{\alpha}$ by 1 and
increasing $oy_{\beta}$ by 1. By Proposition \ref{pr-msetGAC2},
$v=\beta$ is inconsistent. In this case, the values less than
$\beta$ are also inconsistent. Therefore, the smallest value which
is greater than $\beta$ is the tight lower bound. We now need to
show that the conditions of item 3 are exhaustive. If $v=\beta$ is
inconsistent then by Proposition \ref{pr-msetGAC2}, we obtain
$\myvec{ox}>_{lex} \myvec{oy}$ after decreasing $oy_{\alpha}$ by 1
and increasing $oy_{\beta}$ by 1. This is the same as obtaining
$\myvec{ox}>_{lex} \myvec{oy}$ after increasing $ox_{\alpha}$ by 1
and decreasing $ox_{\beta}$ by 1. We have captured this case in the
first two conditions of item 3 in Theorem \ref{th-X3}. Hence, if
$v=\beta$ is inconsistent then we have either
$ox_{\alpha+1}+1=oy_{\alpha}~\And~\gamma~\And~ox_{\beta}>oy_{\beta}+1$
or
$ox_{\alpha+1}+1=oy_{\alpha}~\And~\gamma~\And~ox_{\beta}=oy_{\beta}+1~\And~\sigma$.
\qed

\begin{theorem}
\label{th-Y3} Given $\myvec{ox}=occ(\myfloor(\myvec{X}))$ and
$\myvec{oy}=occ(\myceiling(\myvec{Y}))$ indexed as $u..l$ where
$\myvec{ox} \leq_{lex} \myvec{oy}$, if $max(Y_i) = \alpha$ and
$min(Y_i) > \beta$ then $min(Y_i)$ is the tight lower bound.
\end{theorem}
\proof If $max(Y_i)=\alpha$ then $\alpha \neq -\infty$ and
$\myvec{ox}<_{lex} \myvec{oy}$. Increasing $oy_{min(Y_i)}$ by 1
and decreasing $oy_{\alpha}$ by 1 does not change $\myvec{ox}
<_{lex} \myvec{oy}$. This is because for all
$\alpha>i>max\{l-1,\beta\}$, we have $ox_i \leq oy_i$. Even if now
$\myvec{ox}_{\alpha \rightarrow min(Y_i)+1}=\myvec{oy}_{\alpha
\rightarrow min(Y_i)+1}$, at $min(Y_i)$ we have
$ox_{min(Y_i)}<oy_{min(Y_i)}$. By Proposition \ref{pr-msetGAC2},
$min(Y_i)$ is consistent. \qed

\begin{theorem}
\label{th-Y1}Given $\myvec{ox}=occ(\myfloor(\myvec{X}))$ and
$\myvec{oy}=occ(\myceiling(\myvec{Y}))$ indexed as $u..l$ where
$\myvec{ox} \leq_{lex} \myvec{oy}$, if $max(Y_i)<\alpha$ then
$min(Y_i)$ is the tight lower bound.
\end{theorem}
\proof If $max(Y_i)<\alpha$ then we have $\alpha \neq -\infty$ and
$\myvec{ox} <_{lex} \myvec{oy}$. Decreasing $oy_{max(Y_i)}$ by 1
does not change this. By Proposition \ref{pr-msetGAC2}, $min(Y_i)$
is consistent. \qed

\begin{theorem}
\label{th-Y2}Given $\myvec{ox}=occ(\myfloor(\myvec{X}))$ and
$\myvec{oy}=occ(\myceiling(\myvec{Y}))$ indexed as $u..l$ where
$\myvec{ox} \leq_{lex} \myvec{oy}$, if $max(Y_i) > \alpha$ then
$max(Y_i)$ is the tight lower bound.
\end{theorem}
\proof Decreasing $oy_{max(Y_i)}$ by 1 gives $\myvec{ox}
>_{lex} \myvec{oy}$. By Proposition \ref{pr-msetGAC2}, any $v < max(Y_i)$ in $\domain{D}(Y_i)$ is
inconsistent. \qed

\subsection{Algorithm Details and Theoretical Properties}
\label{ch-gacmset:algorithm-details}

In this subsection, we first explain {\MsetLeq} as well as prove
that it is correct and complete. We then discuss its time
complexity.

\begin{algorithm}[t!]
\begin{footnotesize}
 \SetLine
 \AlgData{$\la X_0, X_1, \ldots, X_{n-1} \ra$, $\la Y_0, Y_1, \ldots, Y_{n-1} \ra$}
 \AlgResult{$occ(\myfloor(\myvec{X}))$ and $occ(\myceiling(\myvec{Y}))$ are initialised, GAC($\myvec{X} \leq_{m} \myvec{Y}$)}
 \numline{1}{0}$l:=min( \msetl \myfloor(\myvec{X}) \msetr \cup \msetl \myfloor(\myvec{Y}) \msetr)$\; 
 \numline{2}{0}$u:=max( \msetl \myceiling(\myvec{X}) \msetr \cup \msetl \myceiling(\myvec{Y}) \msetr)$\; 
 \numline{3}{0}$\myvec{ox}:=occ(\myfloor(\myvec{X}))$\;
 \numline{4}{0}$\myvec{oy}:=occ(\myceiling(\myvec{Y}))$\;
 \numline{5}{0}{\MsetLeq}\;
\caption{\ProcNameSty{Initialise}} \label{msoleq-initialize}
\end{footnotesize}
\end{algorithm}

The algorithm is based on Theorems \ref{th-X3}-\ref{th-Y2}. The
pointers and flags are recomputed every time the algorithm is
called, as maintaining them incrementally in an easy way is not
obvious. Fortunately, incremental maintenance of the occurrence
vectors is trivial. When the minimum value in some $\domain{D}(X_i)$
changes, we update $\myvec{ox}$ by incrementing the entry
corresponding to new $min(X_i)$ by 1, and decrementing the entry
corresponding to old $min(X_i)$ by 1. Similarly, when the maximum
value in some $\domain{D}(Y_i)$ changes, we update $\myvec{oy}$ by
incrementing the entry corresponding to new $max(Y_i)$ by 1, and
decrementing the entry corresponding to old $max(Y_i)$ by 1.

When the constraint is first posted, we need to initialise the
occurrence vectors, and call the filtering algorithm {\MsetLeq} to
establish the generalised arc-consistent state with the initial
values of the occurrence vectors. In Algorithm
\ref{msoleq-initialize}, we show the steps of this initialisation.

\begin{theorem}
\ProcNameSty{\em Initialise} initialises $\myvec{ox}$ and
$\myvec{oy}$ correctly. Then it either establishes failure if
$\myvec{X} \leq_m \myvec{Y}$ is disentailed, or prunes all
inconsistent values from $\myvec{X}$ and $\myvec{Y}$ to ensure
GAC($\myvec{X} \leq_m \myvec{Y}$).
\end{theorem}
\proof \ProcNameSty{Initialise} first computes the most and the
least significant indices of the occurrence vectors as $u$ and $l$
(lines 1 and 2). An occurrence vector $occ(\myvec x)$ associated
with $\myvec x$ is indexed in decreasing order of significance from
$max\msetl \myvec{x} \msetr$ to $min\msetl \myvec{x} \msetr$. Our
occurrence vectors are associated with $\myfloor(\myvec{X})$ and
$\myceiling(\myvec{Y})$ but they are also used for checking support
for $max(X_i)$ and $min(Y_i)$ for all $0 \leq i<n$. We therefore
need to make sure that there are corresponding entries. Also, to be
able to compare two occurrence vectors, they need to start and end
with the occurrence of the same value. Therefore, $u$ is $max(
\msetl \myceiling(\myvec{X}) \msetr \cup \msetl
\myceiling(\myvec{Y}) \msetr)$ and $l$ is $min( \msetl
\myfloor(\myvec{X}) \msetr \cup \msetl \myfloor(\myvec{Y}) \msetr)$.

Using these indices, a pair of vectors $\myvec{ox}$ and $\myvec{oy}$
of length $u-l+1$ are constructed and each entry in these vectors
are set to $0$. Then, $ox_{min(X_i)}$ and $oy_{max(Y_i)}$ are
incremented by 1 for all $0 \leq i <n$. Now, for all $u \geq v \geq
l$, $ox_v$ is the number of occurrences of $v$ in $\msetl
\myfloor(\myvec{X})\msetr$. Similarly, for all $u \geq v \geq l$,
$oy_v$ is the number of occurrences of $v$ in $\msetl
\myceiling(\myvec{Y}) \msetr$. This gives us $
\myvec{ox}=occ(\myfloor(\myvec{X}))$ and
$\myvec{oy}=occ(\myceiling(\myvec{Y}))$ (lines 3 and 4). Finally, in
line 5, \ProcNameSty{Initialise} calls the filtering algorithm
{\MsetLeq} which either establishes failure if $\myvec{X} \leq_m
\myvec{Y}$ is disentailed, or prunes all inconsistent values from
$\myvec{X}$ and $\myvec{Y}$ to ensure GAC($\myvec{X} \leq_m
\myvec{Y}$). \qed


Note that when $\myvec{X} \leq_{m} \myvec{Y}$ is GAC, every value in
$\domain{D}(X_i)$ is supported by $\la min(X_{0}), \ldots,
min(X_{i-1}),\\min(X_{i+1}),\ldots,min(X_{n-1}) \ra$, and $\la
max(Y_0),\ldots,max(Y_{n-1}) \ra$. Similarly, every value in
$\domain{D}(Y_i)$ is supported by $\la min(X_0),\ldots,min(X_{n-1})
\ra$ and $\la max(Y_{0}), \ldots,
max(Y_{i-1}),max(Y_{i+1}),\ldots,max(Y_{n-1}) \ra$. So, {\MsetLeq}
is also called by the event handler\index{Propagation!event!
handler} whenever $min(X_i)$ or $max(Y_i)$ of some $i$ in $[0,n)$
changes.

\begin{algorithm}[t!]
\begin{footnotesize}
 \SetLine
 \AlgData{$\la X_0, X_1, \ldots, X_{n-1} \ra$, $\la Y_0, Y_1, \ldots, Y_{n-1} \ra$}
 \AlgResult{GAC($\myvec{X} \leq_{m} \myvec{Y}$)}
 \numline{A1}{0}\ProcNameSty{SetPointersAndFlags}\;
 \numline{B1}{0}\ForEach{$i \in [0,n)$}{
 \numline{B2}{1}\If{$min(X_i) \neq max(X_i)$}{
 \numline{B3}{2}\lIf{$min(X_i) \geq \alpha$}{\ProcNameSty{setMax($X_i,min(X_i)$)}\;}
 \numline{B4}{2}\If{$max(X_i) \geq \alpha~\And~min(X_i)<\alpha$}{
 \numline{B5}{3}\ProcNameSty{setMax($X_i,\alpha$)}\;
 \numline{B6}{3}\If{$ox_{\alpha}+1=oy_{\alpha}~\And~min(X_i)=\beta~\And~\gamma$}{
 \numline{B7}{4}\eIf{$ox_{\beta}=oy_{\beta}+1$}{
 \numline{B8}{5}\lIf{$\sigma$}{\ProcNameSty{setMax($X_i,\alpha-1$)}\;}
                }{
 \numline{B9}{5}\ProcNameSty{setMax($X_i,\alpha-1$)}\;}}
 \numline{B10}{3}\If{$ox_{\alpha}+1=oy_{\alpha}~\And~min(X_i)<\beta~\And~\gamma$}{
 \numline{B11}{4}\ProcNameSty{setMax($X_i,\alpha-1$)}\;}
                }
}
}
 \numline{C1}{0}\ForEach{$i \in [0,n)$}{
 \numline{C2}{1}\If{$min(Y_i) \neq max(Y_i)$}{
 \numline{C3}{2}\lIf{$max(Y_i) >\alpha$}{\ProcNameSty{setMin($Y_i,max(Y_i)$)}\;}
 \numline{C4}{2}\If{$max(Y_i)=\alpha~\And~min(Y_i) \leq \beta$}{
 \numline{C5}{3}\If{$ox_{\alpha}+1=oy_{\alpha}~\And~\gamma$}{
 \numline{C6}{4}\ProcNameSty{setMin($Y_i,\beta$)}\;
 \numline{C7}{4}\eIf{$ox_{\beta}=oy_{\beta}+1$}{
 \numline{C8}{5}\lIf{$\sigma$}{\ProcNameSty{setMin($Y_i,\beta+1$)}\;}
                }{
 \numline{C9}{5}\ProcNameSty{setMin($Y_i,\beta+1$)}\;}}
               }
 }
} \caption{\MsetLeq} \label{msoleq-propagate}
\end{footnotesize}
\end{algorithm}

In Algorithm \ref{msoleq-propagate}, we show the steps of
{\MsetLeq}. Since $\myvec{ox}$ and $\myvec{oy}$ are maintained
incrementally, the algorithm first sets the pointers and flags in
line {\bf A1} via \ProcNameSty{SetPointersAndFlags} using the
current state of these vectors.

\begin{theorem}
\ProcNameSty{\em SetPointersAndFlags} either sets $\alpha$, $\beta$,
$\gamma$, and $\sigma$ as per their definitions, or establishes
failure as $\myvec{X} \leq_m \myvec{Y}$ is disentailed.
\end{theorem}
\proof Line 2 of \ProcNameSty{SetPointersAndFlags} traverses
$\myvec{ox}$ and $\myvec{oy}$, starting at index $u$, until either
it reaches the end of the vectors (because $\myvec{ox}=\myvec{oy}$),
or it finds an index $i$ where $ox_i \neq oy_i$. In the first case,
$\alpha$ is set to $-\infty$ (line 4) as per Definition
\ref{ch:gacmset-def-alpha}. In the second case, $\alpha$ is set to
$i$ only if $ox_i<oy_i$ (line 5). This is correct by Definition
\ref{ch:gacmset-def-alpha} and means that $\myvec{ox} <_{lex}
\myvec{oy}$. If, however, $ox_i>oy_i$ then we have $\myvec{ox}
>_{lex} \myvec{oy}$. By Proposition \ref{pr-msetGAC1}, $\myvec{X} \leq_m \myvec{Y}$
is disentailed and thus \ProcNameSty{SetPointersAndFlags} terminates
with failure (line 3). This also triggers the filtering algorithm to
fail.

If $\alpha \leq l$ then $\beta$ is set to $-\infty$ (line 6) as per
Definition \ref{ch:gacmset-def-beta}. Otherwise, the vectors are
traversed in lines 9-11, starting at index $\alpha-1$, until either
the end of the vectors are reached (because $\myvec{ox}_{\alpha-1
\rightarrow l} \leq_{lex} \myvec{oy}_{\alpha-1 \rightarrow l}$), or
an index $j$ where $ox_j>oy_j$ is found. In the first case, $\beta$
is set to $-\infty$ (line 12), and in the second case, $\beta$ is
set $j$ (line 13) as per Definition \ref{ch:gacmset-def-beta}.
During this traversal, the Boolean flag $temp$ is set to $true$ iff
$\myvec{ox}_{\alpha-1 \rightarrow max\{l,\beta+1\}} =
\myvec{oy}_{\alpha-1 \rightarrow max\{l,\beta+1\}}$. In lines 14 and
15, $\gamma$ is set to $true$ iff $\beta \neq -\infty$, and either
$\beta=\alpha-1$ or $temp$ is $true$ (because $\myvec{ox}_{\alpha-1
\rightarrow \beta+1} = \myvec{oy}_{\alpha-1 \rightarrow \beta+1}$).
This is correct by Definition \ref{ch:gacmset-def-gamma}.

In line 14, $\sigma$ is initialised to $false$. If $\beta \leq l$
then $\sigma$ remains $false$ (line 16) as per Definition
\ref{ch:gacmset-def-sigma}. Otherwise, the vectors are traversed in
line 18, starting at index $\beta-1$, until either the end of the
vectors are reached (because $\myvec{ox}_{\beta-1 \rightarrow l} =
\myvec{oy}_{\beta-1 \rightarrow l}$), or an index $k$ where $ox_k
\neq oy_k$ is found. In the first case, $\sigma$ remains $false$ as
per Definition \ref{ch:gacmset-def-sigma}. In the second case,
$\sigma$ is set to $true$ only if $ox_k>oy_k$ (line 19). This is
correct by Definition \ref{ch:gacmset-def-sigma} and means that
$\myvec{ox}_{\beta-1 \rightarrow l} >_{lex} \myvec{oy}_{\beta-1
\rightarrow l}$ . If, however, $ox_k<oy_k$ then $\sigma$ remains
$false$ as per Definition \ref{ch:gacmset-def-sigma}. \qed

\begin{procedure}[t!]
\begin{footnotesize}
 \numline{1}{0}$i:=u$\;
 \numline{2}{0}\lWhile{$i \geq l~\And~ox_i=oy_i$}{$i:=i-1$\;}
 \numline{3}{0}\lIf{$i \geq l~\And~ox_i>oy_i$}{fail\;}
 \numline{4}{0}{\bf else} \lIf{$i= l-1$}{$\alpha:=-\infty$\;}
 \numline{5}{0}{\bf else} {$\alpha:=i$\;}
 \numline{6}{0}\lIf{$\alpha \leq l$}{$\beta:=-\infty$\;}
 \numline{7}{0}{\bf else} \If{$\alpha >l$}{
 \numline{8}{1}$j:=\alpha-1$, $temp:=true$\;
 \numline{9}{1}\While{$j \geq l~\And~ox_j \leq oy_j$}{
 \numline{10}{2}\lIf{$ox_j<oy_j$}{$temp:=false$\;}
 \numline{11}{2}$j:=j-1$\;}
 \numline{12}{1}\lIf{$j=l-1$}{$\beta:=-\infty$\;}
 \numline{13}{1}{\bf else} {$\beta:=j$\;}}
 \numline{14}{0}$\gamma:=false$, $\sigma:=false$\;
 \numline{15}{0}\lIf{$\beta \neq -\infty~\And~temp$}{$\gamma:=true$\;}
 \numline{16}{0}\If{$\beta>l$}{
 \numline{17}{1}$k:=\beta-1$\;
 \numline{18}{1}\lWhile{$k \geq l~\And~ox_k=oy_k$}{$k:=k-1$\;}
 \numline{19}{1}\lIf{$k \geq l~\And~ox_k>oy_k$}{$\sigma:=true$\;}
}

\caption{SetPointersAndFlags()}
\end{footnotesize}
\end{procedure}

We now analyse the rest of {\MsetLeq}, where the tight upper bound
for $X_i$ and the tight lower bound for $Y_i$,  for all $0 \leq
i<n$, are sought.

\begin{theorem}
{\MsetLeq} either establishes failure if $\myvec{X} \leq_m
\myvec{Y}$ is disentailed, or prunes all inconsistent values from
$\myvec{X}$ and $\myvec{Y}$ to ensure GAC($\myvec{X} \leq_m
\myvec{Y}$).
\end{theorem}
\proof

If $\myvec{X} \leq_m \myvec{Y}$ is not disentailed then we have
$\myvec{ox} \leq_{lex} \myvec{oy}$ by Proposition \ref{pr-msetGAC1}.
This means that $min(X_i)$ and $max(Y_i)$ for all $0 \leq i<n$ are
consistent by Proposition \ref{pr-msetGAC2}. The algorithm therefore
seeks the tight upper bound for $X_i$ only if $max(X_i)>min(X_i)$
(lines {\bf B2-11}), and similarly the tight lower bound for $Y_i$
only if $min(Y_i)<max(Y_i)$ (lines {\bf C2-9}).

For each $\domain{D}(X_i)$: (1) If $min(X_i) \geq \alpha$ then all
values greater than $min(X_i)$ are pruned, giving $min(X_i)$ as the
tight upper bound (line {\bf B3}). This is correct by Theorem
\ref{th-X2}. (2) If $max(X_i) \geq \alpha~\And~min(X_i)<\alpha$
then:
\begin{itemize}
\item all values greater than $\alpha$ are pruned (line {\bf B5});
\item $\alpha$ is pruned if
$ox_{\alpha}+1=oy_{\alpha}~\And~min(X_i)=\beta~\And~\gamma~\And~ox_{\beta}>oy_{\beta}+1$
(line {\bf B9}), or
$ox_{\alpha}+1=oy_{\alpha}~\And~min(X_i)=\beta~\And~\gamma~\And~ox_{\beta}=oy_{\beta}+1~\And~\sigma$
(line {\bf B8}), or
$ox_{\alpha}+1=oy_{\alpha}~\And~min(X_i)<\beta~\And~\gamma$ (line
{\bf B11}).
\end{itemize}
All the values less than $\alpha$ remain in the domain. By Theorem
\ref{th-X3}, all the inconsistent values are removed. (3) If,
however, $max(X_i)<\alpha$ then $max(X_i)$ is the tight upper bound
by Theorem \ref{th-X1}, and thus no pruning is necessary.

For each $\domain{D}(Y_i)$: (1) If $max(Y_i) > \alpha$ then all
values less than $max(Y_i)$ are pruned, giving $max(Y_i)$ as the
tight lower bound (line {\bf C3}). This is correct by Theorem
\ref{th-Y2}. (2) If $max(Y_i) = \alpha~\And~min(Y_i) \leq \beta$
then:
\begin{itemize}
\item all values less  than $\beta$ are pruned if $ox_{\alpha}+1=oy_{\alpha}~\And~\gamma$ (line {\bf
C6});
\item $\beta$ is pruned if
$ox_{\alpha}+1=oy_{\alpha}~\And~\gamma~\And~ox_{\beta}>oy_{\beta}+1$
(line {\bf C9}) or
$ox_{\alpha}+1=oy_{\alpha}~\And~\gamma~\And~ox_{\beta}=oy_{\beta}+1~\And~\sigma$
(line {\bf C8}).
\end{itemize}
All the values greater than $\beta$ remain in the domain. By Theorem
\ref{th-Y4}, all the inconsistent values are removed. (3) If,
however, $max(Y_i)=\alpha~\And~min(Y_i)>\beta$ or $max(Y_i)<\alpha$
then $min(Y_i)$ is the tight lower bound by Theorems \ref{th-Y3} and
\ref{th-Y1}, and thus no pruning is needed.

{\MsetLeq} is a correct and complete filtering algorithm, as it
either establishes failure if $\myvec{X} \leq_m \myvec{Y}$ is
disentailed, or prunes all inconsistent values from $\myvec{X}$ and
$\myvec{Y}$ to ensure GAC($\myvec{X} \leq_m \myvec{Y}$). \qed

When we prune a value, we do not need to check recursively that
previous support remains. The algorithm tightens $max(X_i)$ and
$min(Y_i)$ without touching $min(X_i)$ and $max(Y_i)$, for all $0
\leq i <n$, which provide support for the values in the vectors.
The exception is if a domain wipe out occurs. As the constraint is
not disentailed, we have $\myvec{ox} \leq_{lex} \myvec{oy}$. This
means $min(X_i)$ and $max(Y_i)$ for all $0 \leq i <n$ are
supported. Hence, the prunings of the algorithm cannot cause any
domain wipe-out.

The algorithm works also when the vectors are of different length
as we build and reason about the occurrence vectors as opposed to
the original vectors. Also, we do not assume that the original
vectors are of the same length when we set the pointer $\beta$.

The algorithm corrects a mistake that appears in
\cite{fhkmw:ijcai03}. We have noticed that in \cite{fhkmw:ijcai03}
we do not always prune the values greater than $\alpha$ when we have
$max(X_i) \geq \alpha$ and $min(X_i)<\alpha$. As shown above, this
algorithm is correct and complete.

To improve the time complexity,
we assume that domains are transformed
so that their union is a continuous
interval. Suppose, for instance, that we have variables
with domains $\{1,5\}$, $\{1,100\}$ and $\{5,100\}$. This
transformation normalises the domains to
$\{1,2\}$, $\{1,3\}$ and $\{2,3\}$. This
This technique is widely used (see for instance \cite{gcc:bc3}) and
does not change the worst-case complexity of our propagator.
It gives us a tighter upper bound on the
complexity of our propagator in terms of the number
of distinct values as compared to the difference between
the largest and smallest values.

%
%

\begin{theorem}
\ProcNameSty{\em Initialise} runs in time $O(n+d)$, where $d$ is
the number of distinct values.
\end{theorem}
\proof \ProcNameSty{Initialise} first constructs $\myvec{ox}$ and
$\myvec{oy}$ of length $d$ where each entry is zero, and then
increments $ox_{min(X_i)}$ and $oy_{max(Y_i)}$ by 1 for all $0
\leq i <n$. Hence, the complexity of initialisation is $O(n+d)$.
\qed

\begin{theorem}
{\MsetLeq} runs in time $O(nb+d)$, where $b$ is the cost of
adjusting the bounds of a variable, and $d$ is the number of different
values.
\end{theorem}
\proof {\MsetLeq} does not construct $\myvec{ox}$ and
$\myvec{oy}$, but rather uses their most up-to-date states.
{\MsetLeq} first sets the pointers and flags which are defined on
$\myvec{ox}$ and $\myvec{oy}$. In the worst case both vectors are
traversed once from the beginning until the end, which gives an
$O(d)$ complexity. Next, the algorithm goes through every variable
in the original vectors $\myvec{X}$ and $\myvec{Y}$ to check for
support. Deciding the tight bound for each variable is a constant
time operation, but the cost of adjusting the bound is $b$. Since
we have $O(n)$ variables, the complexity of the algorithm is
$O(nb+d)$. \qed


If $d \ll n$ then the algorithm runs in time $O(nb)$. Since a
multiset is a set with possible repetitions, we expect that the
number of distinct values in a multiset is often less than the
cardinality of the multiset, giving us a linear time filtering
algorithm.

\section{Multiset Ordering with Large Domains}
\label{ch-gacmset:algo2} {\MsetLeq} is a linear time algorithm in
the $n$ given that $d \ll n$. If
instead we have $n \ll d$ then the complexity of the algorithm is
$O(d)$, dominated by the cost of the construction of the
occurrence vectors and the initialisation of the pointers and flags.
This can happen, for instance, when the vectors being multiset
ordered are variables in the occurrence representation of a multiset
\cite{kw:symcon02}. Is there then an alternative way of propagating
the multiset ordering constraint whose complexity is independent of
the domains?

\subsection{Remedy}
In case $d$ is a large number, it could be costly to construct the
occurrence vectors. We can instead sort $\myfloor(\myvec{X})$ and
$\myceiling(\myvec{Y})$, and compute $\alpha$, $\beta$, $\gamma$,
$\sigma$, and the number of occurrences of $\alpha$ and $\beta$ in
$\msetl \myfloor(\myvec{X}) \msetr$ and $\msetl
\myceiling(\myvec{Y}) \msetr$ as if we had the occurrence vectors by
scanning these sorted vectors. This information is all we need to
find support for the bounds of the variables. Let us illustrate this
on an example. To simplify presentation, we assume that the vectors
are of the same length. Consider $\myvec{X} \leq_m \myvec{Y}$ where
$\myvec{sx}=sort(\myfloor(\myvec{X}))$ and
$\myvec{sy}=sort(\myceiling(\myvec{Y}))$ are as follows:
\[
\begin{array}{cccccccccccc}
 \myvec{sx} &=& \la 5, & 4, & 3, &2, &2, &2, &2, &1 \ra\\
 \myvec{sy} &=& \la 5, & 4, & 4, &4, &3, &1, &1, &1 \ra
\end{array}
\]
We traverse $\myvec{sx}$ and $\myvec{sy}$ until we find an index
$i$ such that $sx_i<sy_i$, and for all $0 \leq t < i$ we have
$sx_t=sy_t$. In our example, $i$ is $2$:
\[
\begin{array}{cccccccccccc}
  &&       &     & \downarrow i &&&&& \\
 \myvec{sx} &=& \la 5, & 4, & 3, &2, &2, &2, &2, &1 \ra\\
 \myvec{sy} &=& \la 5, & 4, & 4, &4, &3, &1, &1, &1 \ra

 \end{array}
\]
This means that the number occurrences of any value greater than
$sy_i$ are equal in $\msetl \myfloor(\myvec{X}) \msetr$ and in
$\msetl \myceiling(\myvec{Y}) \msetr$, but there are more occurrence
of $sy_i$ in $\msetl \myceiling(\myvec{Y}) \msetr$ than in $\msetl
\myfloor(\myvec{X}) \msetr$. That is, $ox_5=oy_5$ and $ox_4<oy_4$.
By Definition \ref{ch:gacmset-def-alpha}, $\alpha$ is equal to $4$.
We now move only along $\myvec{sy}$ until we find an index $j$ such
that $sy_j \neq sy_{j-1}$, so that we reason about the number of
occurrences of the smaller values. In our example, $j$ is $4$:
\[
\begin{array}{cccccccccccc}
            &&       &     & \downarrow i &&&&&\\
 \myvec{sx} &=& \la 5, & 4, & 3, &2, &2, &2, &2, &1 \ra\\
 \myvec{sy} &=& \la 5, & 4, & 4, &4, &3, &1, &1, &1 \ra \\
             &&       &     &&&\uparrow j&&&
 \end{array}
\]
We here initialise $\gamma$ to $true$, and start traversing
$\myvec{sx}$ and $\myvec{sy}$ simultaneously. We have
$sx_i=sy_j=3$. This adds $1$ to $ox_3$ and $oy_3$, keeping
$\gamma=true$. We move one index ahead in both vectors by
incrementing $i$ to $3$ and $j$ to $5$:
\[
\begin{array}{cccccccccccc}
             &&       &     &  & \downarrow i &&&&\\
 \myvec{sx} &=& \la 5, & 4, & 3, &2, &2, &2, &2, &1 \ra\\
 \myvec{sy} &=& \la 5, & 4, & 4, &4, &3, &1, &1, &1 \ra \\
             &&       &     & && & \uparrow j&&
 \end{array}
\]
We now have $sx_i>sy_j$, which suggests that $sx_i$ occurs at
least once in $\msetl \myfloor(\myvec{X}) \msetr$ but does not
occur in $\msetl \myceiling(\myvec{Y}) \msetr$. That is, $ox_2>0$
and $oy_2=0$. By Definition \ref{ch:gacmset-def-beta}, $\beta$
points to $2$. This does not change that $\gamma$ is $true$. We
now move only along $\myvec{sx}$ by incrementing $i$ until we find
$sx_i \neq sx_{i-1}$, so that we reason about the number of
occurrences of the smaller values:
\[
\begin{array}{cccccccccccc}
            &&       &     & &&&&&\downarrow i \\
 \myvec{sx} &=& \la 5, & 4, & 3, &2, &2, &2, &2, &1 \ra\\
 \myvec{sy} &=& \la 5, & 4, & 4, &4, &3, &1, &1, &1 \ra \\
             &&       &     &&&&\uparrow j&&
 \end{array}
\]
With the new value of $i$, we have $sx_i=sy_i=1$. This increases
both $ox_1$ and $oy_1$ by one. Reaching the end of only
$\myvec{sx}$ hints the following: either $1$ occurs more than once
in $\msetl \myceiling(\myvec{Y}) \msetr$, or it occurs once but
there are values in $\msetl \myceiling(\myvec{Y}) \msetr$ less
than $1$ and they do not occur in $\msetl \myfloor(\myvec{X})
\msetr$. By Definition \ref{ch:gacmset-def-sigma}, $\gamma$ is
$false$.

Finally, we need to know the number of occurrences of $\alpha$ and
$\beta$ in $\msetl \myfloor(\myvec{X}) \msetr$ and $\msetl
\myceiling(\myvec{Y}) \msetr$. Since we already know what $\alpha$
and $\beta$ are, another scan of $\myvec{sx}$ and $\myvec{sy}$
gives us the needed information: for all $0 \leq i <n$, we
increment $ox_\alpha$ (resp. $ox_\beta$) by 1 if $sx_i=\alpha$
(resp. $sx_i=\beta$), and also $oy_\alpha$ (resp. $oy_\beta$) by 1
if $sy_i=\alpha$ (resp. $sy_i=\beta$).

\subsection{An Alternative Filtering Algorithm}
As witnessed in the previous section, it suffices to sort
$\myfloor(\myvec{X})$ and $\myceiling(\myvec{Y})$, and scan the
sorted vectors to compute $\alpha$\index{alpha@$\alpha$},
$\beta$\index{beta@$\beta$}, $\gamma$\index{gamma@$\gamma$},
$\sigma$\index{sigma@$\sigma$}, $ox_\alpha$, $oy_\alpha$,
$ox_\beta$, and $oy_\beta$. We can then directly reuse lines {\bf
B1-11} and {\bf C1-9} of {\MsetLeq} to obtain a new filtering
algorithm\index{Filtering algorithm}. As a result, we need to
change only \ProcNameSty{Initialise} and
\ProcNameSty{SetPointersAndFlags}.

\begin{algorithm}[t!]
\begin{footnotesize}
 \SetLine
 \AlgData{$\la X_0, X_1, \ldots, X_{n-1} \ra$, $\la Y_0, Y_1, \ldots, Y_{n-1} \ra$}
 \AlgResult{$sort(\myfloor(\myvec{X}))$ and $sort(\myceiling(\myvec{Y}))$ are initialised, GAC($\myvec{X} \leq_{m} \myvec{Y}$)}

\numline{1}{0}$\myvec{sx}:=sort(\myfloor(\myvec{X}))$\;
 \numline{2}{0}$\myvec{sy}:=sort(\myceiling(\myvec{Y}))$\;
 \numline{3}{0}{\MsetLeq}\;
\caption{\ProcNameSty{Initialise}} \label{msoleq-initialize-new}
\end{footnotesize}
\end{algorithm}

In Algorithm \ref{msoleq-initialize-new}, we show the new
\ProcNameSty{Initialise}. Instead of constructing a pair of
occurrence vectors associated with $\myfloor(\myvec{X})$ and
$\myceiling(\myvec{Y})$, we now sort $\myfloor(\myvec{X})$ and
$\myceiling(\myvec{Y})$ and then call {\MsetLeq}.

Similar to the original algorithm, we recompute the pointers and
flags every time we call the filtering algorithm. Maintaining the
sorted vectors incrementally is trivial. When the minimum value in
some $\domain{D}(X_i)$ changes, we update $\myvec{sx}$ by inserting
the new $min(X_i)$ into, and removing the old $min(X_i)$ from
$\myvec{sx}$. Similarly, when the maximum value in some
$\domain{D}(Y_i)$ changes, we update $\myvec{sy}$ by inserting the
new $max(Y_i)$ into, and removing the old $max(Y_i)$ from
$\myvec{sy}$. Since these vectors need to remain sorted after the
update, such modifications require binary search. The cost of
incrementality thus increases  from $O(1)$ to $O(log(n))$ compared
to the original filtering algorithm.


\begin{procedure}[t!]
\begin{footnotesize}
 \numline{1}{0}$i:=0$\;
 \numline{2}{0}\lWhile{$i <n~\And~sx_i=sy_i$}{$i:=i+1$\;}
 \numline{3}{0}\lIf{$i < n~\And~sx_i>sy_i$}{fail\;}
 \numline{4}{0}{\bf else} \lIf{$i=n$}{$\alpha:=-\infty$, $\beta:=-\infty$, $\gamma:=false$, $\sigma:=false$, return\;}
 \numline{5}{0}{\bf else} {$\alpha:=sy_i$\;}
 \numline{6}{0}$\gamma:=true$\;
 \numline{7}{0}$j:=i+1$\;
 \numline{8}{0}\lWhile{$j<n~\And~sy_j=sy_{j-1}$}{$j:=j+1$\;}
 \numline{9}{0}\lIf{$j=n$}{$\beta:=sx_i$\;}
 \numline{10}{0}{\bf else} \If{$j<n$}{
 \numline{11}{1}\While{$i<n~\And~j<n$}{
 \numline{12}{2}\lIf{$sx_i>sy_j$}{$\beta:=sx_i$, {\bf break}\;}
 \numline{13}{2}\lIf{$sx_i<sy_j$}{$\gamma:=false$, $j:=j+1$\;}
 \numline{14}{2}\lIf{$sx_i=sy_j$}{$i:=i+1$, $j:=j+1$\;}}
 \numline{15}{1}\lIf{$j=n$}{$\beta:=sx_i$\;}}
 \numline{16}{0}$k:=i+1$\;
 \numline{17}{0}\lWhile{$k <n~\And~sx_k=sx_{k-1}$}{$k:=k+1$\;}
 \numline{18}{0}\lIf{$k=n$}{$\sigma:=false$\;}
 \numline{19}{0}{\bf else} \If{$k<n$}{
 \numline{20}{1}\While{$k<n~\And~j<n$}{
 \numline{21}{2}\lIf{$sx_k>sy_j$}{$\sigma:=true$, {\bf break}\;}
 \numline{22}{2}\lIf{$sx_k<sy_j$}{$\sigma:=false$, {\bf break}\;}
 \numline{23}{2}\lIf{$sx_k=sy_j$}{$k:=k+1$, $j:=j+1$\;}}
 \numline{24}{1}\lIf{$k=n$}{$\sigma:=false$ \;}
 \numline{25}{1}{\bf else} \If{$j=n$}{$\sigma:=true$\;}}
 \numline{26}{0}$i:=0$, $ox_{\alpha}=0$, $oy_{\alpha}=0$, $ox_{\beta}=0$, $oy_{\beta}=0$\;
 \numline{27}{0}\ForEach{$i \in [0,n)$}{
 \numline{28}{1}\lIf{$sx_i=\alpha$}{$ox_{\alpha}:=ox_{\alpha}+1$\;}
 \numline{29}{1}\lIf{$sx_i=\beta$}{$ox_{\beta}:=ox_{\beta}+1$\;}
 \numline{30}{1}\lIf{$sy_i=\alpha$}{$oy_{\alpha}:=oy_{\alpha}+1$\;}
 \numline{31}{1}\lIf{$sy_i=\beta$}{$oy_{\beta}:=oy_{\beta}+1$\;}}

\caption{SetPointersAndFlags()}
\end{footnotesize}
\end{procedure}

Given the most up-to-date $\myvec{sx}$ and $\myvec{sy}$, how do we
set our pointers and flags? In line 2 of our new
\ProcNameSty{SetPointersAndFlags}, we traverse $\myvec{sx}$ and
$\myvec{sy}$, starting at index 0, until either we reach the end
of the vectors (because the vectors are equal), or we find an
index $i$ where $sx_i \neq sy_i$. In the first case, we first set
$\alpha$ and $\beta$ to $-\infty$, and $\gamma$ and $\sigma$ to
$false$, and then return (line 4). In the second case, if
$sx_i>sy_i$ then disentailment is detected and
\ProcNameSty{SetPointersAndFlags} terminates with failure (line
3). The reason of the return and failure is due to the following
theoretical result.

\begin{theorem}
\label{th-sort} $occ(\myvec{x}) \leq_{lex} occ(\myvec{y})$ iff
$sort(\myvec{x}) \leq_{lex} sort(\myvec{y})$.
\end{theorem}
\proof ($\Rightarrow$) If $occ(\myvec{x}) <_{lex} occ(\myvec{y})$
then a value  $a$ occurs more in $\msetl \myvec{y} \msetr$ than in
$\msetl \myvec{x} \msetr$, and the occurrence of any value $b>a$
is the same in both multisets. By deleting all the occurrences of
$a$ from $\msetl \myvec{x} \msetr$ and the same number of
occurrences of $a$ from $\msetl \myvec{y} \msetr$, as well as any
$b>a$ from both multisets, we get $max\msetl \myvec{x}
\msetr<max\msetl \myvec{y} \msetr$. Since the leftmost values in
$sort(\myvec{x})$ and $ sort(\myvec{y})$ are $max\msetl \myvec{x}
\msetr$ and $max\msetl \myvec{y} \msetr$ respectively, we have
$sort(\myvec{x}) <_{lex} sort(\myvec{y})$. If $occ(\myvec{x})=
occ(\myvec{y})$ then we have $\msetl \myvec{x} \msetr =\msetl
\myvec{y} \msetr$. By sorting the elements in $\myvec{x}$ and
$\myvec{y}$, we obtain the same vectors. Hence, $sort(\myvec{x}) =
sort(\myvec{y})$.

($\Leftarrow$) Suppose $\myvec{ox}=occ(\myvec{x})$,
$\myvec{oy}=occ(\myvec{y})$, $\myvec{sx}=sort(\myvec{x})$, $
\myvec{sy}=sort(\myvec{y})$, and we have $\myvec{sx}= \myvec{sy}$.
Then $\msetl \myvec{x} \msetr$ and $\msetl \myvec{y} \msetr$
contain the same elements with equal occurrences. Hence,
$\myvec{ox}=\myvec{oy}$. Suppose $\myvec{sx} <_{lex} \myvec{sy}$.
If $sx_0<sy_0$ then the leftmost index of $\myvec{ox}$ and
$\myvec{oy}$ is $sy_0$, and we have $ox_{sy_0}=0$ and
$oy_{sy_0}>0$. This gives $\myvec{ox} <_{lex} \myvec{oy}$. If
$sx_0=sy_0=a$ then we eliminate one occurrence of $a$ from $\msetl
\myvec{x} \msetr$ and $\msetl \myvec{y} \msetr$, and compare the
resulting multisets. \qed

Hence,  whenever we have $\myvec{sx} \geq_{lex} \myvec{sy}$, we
proceed as if we had $occ(\myfloor(\myvec{X})) \geq_{lex}
occ(\myceiling (\myvec{Y}))$. But then what do we do if we have
$\myvec{sx} <_{lex} \myvec{sy}$? In line 5, we have $sx_i<sy_i$ and
$sx_t=sy_t$ for all $0 \leq t <i$. This means that the number
occurrences of any value greater than $sy_i$ are equal in $\msetl
\myfloor(\myvec{X}) \msetr$ and in $\msetl \myceiling(\myvec{Y})
\msetr$, but there are more occurrence of $sy_i$ in $\msetl
\myceiling(\myvec{Y}) \msetr$ than in $\msetl \myfloor(\myvec{X})
\msetr$. Therefore, we here set $\alpha$ to $sy_i$.

After initialising $\gamma$ to $true$ in line 6, we start seeking a
value for $\beta$. For the sake of simplicity, we here assume our
original vectors are of same length. Hence, $\beta$ cannot be
$-\infty$ as $\alpha$ is not $-\infty$. In line 8, we traverse
$\myvec{sy}$, starting at index $i+1$, until either we reach the end
of the vector (because all the remaining values in $\msetl
\myceiling(\myvec{Y}) \msetr$ are $sy_i$), or we find an index $j$
such that $sy_j \neq sy_{j-1}$. In the first case, we set $\beta$ to
$sx_i$ (line 9) because $sx_i$ occurs at least once in $\msetl
\myfloor(\myvec{X}) \msetr$ but does not occur in $\msetl
\myceiling(\myvec{Y}) \msetr$. Since no value between $\alpha$ and
$\beta$ occur more in $\msetl \myceiling(\myvec{Y}) \msetr$ than in
$\msetl \myfloor(\myvec{X}) \msetr$, $\gamma$ remains $true$. In the
second case, $sy_j$ gives us the next largest value in $\msetl
\myceiling(\myvec{Y}) \msetr$. In lines 11-14, we traverse
$\myvec{sx}$ starting from $i$, and $\myvec{sy}$ starting from $j$.
If $sx_i>sy_j$ then we set $\beta$ to $sx_i$ (line 12) because
$sx_i$ occurs more in $\msetl \myfloor(\myvec{X}) \msetr$ than in
$\msetl \myceiling(\myvec{Y}) \msetr$. Having found the value of
$\beta$, we here exit the while loop using {\bf break}. If
$sx_i<sy_j$ then $sy_j$ occurs more in $\msetl \myceiling(\myvec{Y})
\msetr$ than in $\msetl \myfloor(\myvec{X}) \msetr$. Since we are
still looking for a value for $\beta$, we set $\gamma$ to $false$
(line 13). We then move to the next index in $\myvec{sy}$ to find
the next largest value in $\msetl \myceiling(\myvec{Y}) \msetr$. If
$sx_i=sy_j$ then we move to the next index both in $\myvec{sx}$ and
$\myvec{sy}$ to find the next largest values in $\msetl
\myfloor(\myvec{X}) \msetr$ and $\msetl \myceiling(\myvec{Y})
\msetr$ (line 14). As $j$ is at least one index ahead of $i$, $j$
can reach to $n$ before $i$ does during this traversal. In such a
case, we set $\beta$ to $sx_i$ (line 15) due to the same reasoning
as in line 12.

The process of finding the value of $\sigma$ (lines 16-25) is very
similar to that of $\beta$. In line 17, we traverse $\myvec{sx}$,
starting at index $i+1$, until either we reach the end of the
vector (because all the remaining values in $\msetl
\myfloor(\myvec{X}) \msetr$ are $\beta$), or we find an index $k$
such that $sx_k \neq sx_{k-1}$. In the first case, we set $\sigma$
to $false$ (line 18) because either $sy_j$ occurs at least once in
$\msetl \myceiling(\myvec{Y}) \msetr$ but does not occur in
$\msetl \myfloor(\myvec{X}) \msetr$ (due to line 12), or there are
no values less than $\beta$ both in  $\msetl \myfloor(\myvec{X})
\msetr$ and in $\msetl \myceiling(\myvec{Y}) \msetr$ (due to line
15). In the second case, $sx_k$ gives us the next largest value in
$\msetl \myfloor(\myvec{X}) \msetr$. In lines 20-23, we traverse
$\myvec{sx}$ starting from $k$, and $\myvec{sy}$ starting from
$j$. The reasoning now is very similar to that of the traversal
for $\beta$. Instead of setting a value for $\beta$, we set
$\sigma$ to $true$, and instead of setting $\gamma$ to $false$, we
set $\sigma$ to $false$, for the same reasons. If $k$ reaches $n$
before $j$, then we set $\sigma$ to $false$ (line 24) due to the
same reason as in line 22. If $k$ and $j$ reach $n$ together, then
again we set $\sigma$ to $false$, because we have the same number
of occurrences of any value less than $\beta$ in $\msetl
\myfloor(\myvec{X}) \msetr$ and in $\msetl \myceiling(\myvec{Y})
\msetr$. If, however, $j$ reaches $n$ before $k$, then we set
$\sigma$ to $true$ (line 25) due to the same reason as in line 21.

Finally, we go through each of $sx_i$ and $sy_i$ in lines 26-31,
and find how many times $\alpha$ and $\beta$ occur in $\msetl
\myfloor(\myvec{X}) \msetr$ and in $\msetl \myceiling(\myvec{Y})
\msetr$, by counting how many times $\alpha$ and $\beta$ occur in
$\myvec{sx}$ and in $\myvec{sy}$, respectively.

The complexity of this new algorithm is independent of the domains
and is $O(n~log(n))$, as the cost of sorting dominates.

\section{Extensions}
\label{ch-gacmset:extensions} In this section, we answer two
important questions. First, how can we enforce strict multiset
ordering? Second, how can we detect entailment?

\subsection{Strict Multiset Ordering Constraint}
\label{ch-gacmset:strictmset} We can easily get a filtering
algorithm\index{Filtering algorithm} for strict multiset ordering
constraint by slightly modifying {\MsetLeq}. This new algorithm,
called {\MsetLess}, either detects the disentailment of $\myvec{X}
<_m \myvec{Y}$, or prunes inconsistent values to perform GAC on
$\myvec{X} <_m \myvec{Y}$.  Before showing how we modify {\MsetLeq},
we first study $\myvec{X} <_m \myvec{Y}$ from a theoretical point of
view. It is not difficult to modify Theorems \ref{th-msetGAC1},
\ref{th-msetGAC2} and \ref{th-occ} so as to exclude the equality and
obtain the following propositions:

\begin{proposition}
\label{pr-mset-strict-GAC1} $\myvec{X} <_m \myvec{Y}$ is
disentailed\index{Disentailment} iff $occ( \myfloor(\myvec{X}))
\geq_{lex} occ(\myceiling(\myvec{Y}))$.
\end{proposition}

\begin{proposition}
\label{pr-mset-strict-GAC2} GAC($\myvec{X} <_m
\myvec{Y}$)\index{Arc-consistency!generalised} iff for all $i$ in
$[0,n)$:
\begin{eqnarray*}
 occ(\myfloor(\myvec{X}_{X_i \assigned max(X_i)})) &<_{lex}& occ( \myceiling(\myvec{Y}))\\
 occ(\myfloor(\myvec{X})) & <_{lex}& occ(\myceiling(\myvec{Y}_{Y_i \assigned
 min(Y_i)}))
\end{eqnarray*}
\end{proposition}
We can exploit the similarity between Proposition
\ref{pr-msetGAC2} and \ref{pr-mset-strict-GAC2}, and find the
tight consistent bounds by making use of the occurrence vectors
$\myvec{ox}=occ(\myfloor(\myvec{X}))$ and
$\myvec{oy}=occ(\myceiling(\myvec{Y}))$, the pointers, and the
flags. In Theorems \ref{th-X3} to \ref{th-Y2}, we have $\myvec{ox}
\leq_{lex} \myvec{oy}$. We decide whether a value $v$ in some
domain $D$ is consistent or not by first increasing $ox_v$/$oy_v$
by 1, and then decreasing $min(D)$/$max(D)$ by 1. The value is
consistent for $\myvec{X} \leq_m \myvec{Y}$ iff the change gives
$\myvec{ox} \leq_{lex} \myvec{oy}$. In Theorems \ref{th-X2} and
\ref{th-Y2}, changing the occurrences gives  $\myvec{ox}
>_{lex} \myvec{oy}$. This means that $v$ is
inconsistent not only for $\myvec{X} \leq_m \myvec{Y}$ but also
for $\myvec{X} <_m \myvec{Y}$. In Theorems \ref{th-X1},
\ref{th-Y3}, and \ref{th-Y1}, however, we initially have
$\myvec{ox} <_{lex} \myvec{oy}$ and  changing the occurrences does
not disturb the strict lexicographic ordering. This suggests $v$
is consistent also for  $\myvec{X} <_m \myvec{Y}$.

In Theorems \ref{th-X3} and \ref{th-Y4}, we initially have
$\myvec{ox}<_{lex} \myvec{oy}$, and after the change we obtain
either of $\myvec{ox}>_{lex} \myvec{oy}$, $\myvec{ox} =
\myvec{oy}$, and $\myvec{ox} <_{lex} \myvec{oy}$. In the first
case $v$ is inconsistent, whereas in the third case $v$ is
consistent, for both constraints. In the second case, however, $v$
is consistent for $\myvec{X} \leq_m \myvec{Y}$ but not for
$\myvec{X} <_m \myvec{Y}$. This case arises if we get
$\myvec{ox}_{u \rightarrow \beta}=\myvec{oy}_{u \rightarrow
\beta}$ by the change to the occurrence vectors, and we have
either $\beta>l$ and $\myvec{ox}_{\beta-1 \rightarrow
l}=\myvec{oy}_{\beta-1 \rightarrow l}$, or $\beta=l$. We therefore
need to record whether there are any subvectors below $\beta$, and
if this is the case we need to know whether they are equal. This
can easily be done by extending the definition of $\sigma$ which
already tells us whether we have $\beta>l$ and
$\myvec{ox}_{\beta-1 \rightarrow l}>_{lex}\myvec{oy}_{\beta-1
\rightarrow l}$.
\begin{definition}
\label{ch:gacmset-def-sigma-new}Given
$\myvec{ox}=occ(\myfloor(\myvec{X}))$ and
$\myvec{oy}=occ(\myceiling(\myvec{Y}))$ indexed as $u..l$ where
$\myvec{ox} <_{lex} \myvec{oy}$, the flag
$\sigma$\index{sigma@$\sigma$} is $true$ iff:
\[
(\beta > l ~\And~\myvec{ox}_{\beta-1 \rightarrow l} \geq_{lex}
\myvec{oy}_{\beta-1 \rightarrow l}) ~\Or~\beta=l
\]
\end{definition}
Theorems \ref{th-X3} and \ref{th-Y4} now declare a value
inconsistent if we get $\myvec{ox}_{u \rightarrow
\beta}=\myvec{oy}_{u \rightarrow \beta}$ 
when the occurrence vectors change, and we have either $\beta>l$
and $\myvec{ox}_{\beta-1 \rightarrow l}=\myvec{oy}_{\beta-1
\rightarrow l}$, or $\beta=l$.

%
%

How do we now modify {\MsetLeq} to obtain the filtering algorithm
{\MsetLess}? Theorems \ref{th-X1}, \ref{th-X2}, \ref{th-Y3},
\ref{th-Y1}, and \ref{th-Y2} are valid also for $\myvec{X} <_m
\myvec{Y}$. Moreover, Theorems \ref{th-X3} and \ref{th-Y4} can
easily be adapted for $\myvec{X} <_m \myvec{Y}$ by changing the
definition of $\sigma$. Hence, the pruning part of the algorithm
need not to be modified, provided that $\sigma$ is set correctly.
Also, by Proposition \ref{pr-mset-strict-GAC1}, we need to fail
under the new disentailment condition. These suggest we only need
to revise \ProcNameSty{SetPointersAndFlags}, so that we fail
whenever we have $\myvec{ox} \geq_{lex} \myvec{oy}$, and set
$\sigma$ to $true$ also when we have $\beta=l$, or $\beta>l$ and
$\myvec{ox}_{\beta-1 \rightarrow l}=\myvec{ox}_{\beta-1
\rightarrow l}$. This corrects a mistake in \cite{fhkmw:ijcai03}
which claims that failing whenever we have  $\myvec{ox} \geq_{lex}
\myvec{oy}$ and setting $\beta$ to $l-1$ as opposed to $-\infty$
are enough to achieve strict multiset ordering.

\subsection{Entailment}
\label{ch-gacmset:entailment}

%
{\MsetLeq} is a correct and complete filtering algorithm. However,
it does not detect entailment. Even though detecting entailment
does not change the semantics of the algorithm, it can lead to
significant savings from an operational point of view.
%
We thus introduce another Boolean flag, called $entailed$, which
indicates whether $\myvec{X} \leq_m \myvec{Y}$ is entailed. More
formally:

\begin{definition}
\label{ch:gacmset-def-entailed} Given $\myvec{X}$ and $\myvec{Y}$,
the flag {\em entailed}\index{entailed@$entailed$} is set to
$true$ iff $\myvec{X} \leq_{m} \myvec{Y}$ is $true$.
\end{definition}


The multiset ordering constraint is entailed whenever the largest
value that $\myvec{X}$ can take is less than or equal to the
smallest value that $\myvec{Y}$ can take under the ordering in
concern.

\begin{theorem}
\label{th-msetGAC3} $\myvec{X} \leq_m \myvec{Y}$ is
entailed\index{Entailment} iff $\msetl \myceiling(\myvec{X})
\msetr \leq_m \msetl \myfloor(\myvec{Y} )\msetr$.
\end{theorem}
\proof ($\Rightarrow$) Since  $\myvec{X} \leq_m \myvec{Y}$ is
entailed, any combination of assignments, including $\myvec{X}
\assigned \myceiling(\myvec{X})$ and $\myvec{Y} \assigned
\myfloor(\myvec{Y})$, satisfies $\myvec{X} \leq_{m} \myvec{Y}$.
Hence, $\msetl \myceiling(\myvec{X}) \msetr \leq_m \msetl
\myfloor(\myvec{Y} )\msetr$.

($\Leftarrow$) Any $\myvec{x}\in \myvec{X}$ is less  than or equal
to any $\myvec{y}\in \myvec{Y}$ under multiset ordering. Hence,
$\myvec{X} \leq_m \myvec{Y}$ is entailed. \qed

By Theorems \ref{th-occ} and \ref{th-msetGAC3}, we can detect
entailment by lexicographically comparing the occurrence vectors
associated with $\myceiling(\myvec{X})$ and $\myfloor(\myvec{Y}
)$.
\begin{proposition}
\label{pr-msetGAC3} $\myvec{X} \leq_m \myvec{Y}$ is
entailed\index{Entailment} iff
$occ(\myceiling(\myvec{X}))\leq_{lex} occ(\myfloor(\myvec{Y}))$.
\end{proposition}

When {\MsetLeq} is executed, we have three possible scenarios in
terms of entailment: (1) $\myvec{X} \leq_m \myvec{Y}$ has already
been entailed in the past due to the previous modifications to the
variables; (2) $\myvec{X} \leq_m \myvec{Y}$ was not entailed
before, but after the recent modifications which invoked the
algorithm, $\myvec{X} \leq_m \myvec{Y}$ is now entailed; (3)
$\myvec{X} \leq_m \myvec{Y}$ has not been entailed, but after the
prunings of the algorithm, $\myvec{X} \leq_m \myvec{Y}$ is now
entailed. In all cases, we can safely return from the algorithm.
We need to, however, record entailment in our flag $entailed$ in
the second and the third cases, before returning.

\begin{algorithm}[t!]
\begin{footnotesize}
 \SetLine
 \AlgData{$\la X_0, X_1, \ldots, X_{n-1} \ra$, $\la Y_0, Y_1, \ldots, Y_{n-1} \ra$}
 \AlgResult{$occ(\myfloor(\myvec{X}))$, $occ(\myceiling(\myvec{Y}))$, $occ(\myceiling(\myvec{X}))$, $occ(\myfloor(\myvec{Y}))$, and $entailed$ are initialised, GAC($\myvec{X} \leq_{m} \myvec{Y}$)}
 \numline{0}{0}$entailed:=false$\;
 \vdots
 \numline{5}{0}$\myvec{ex}:=occ(\myceiling(\myvec{X}))$\;
 \numline{6}{0}$\myvec{ey}:=occ(\myfloor(\myvec{Y}))$\;
 \numline{7}{0}{\MsetLeq}\;
\caption{\ProcNameSty{Initialise}}
\label{msoleq-initialize-entailed}
\end{footnotesize}
\end{algorithm}

To deal with entailment, we need to modify both
\ProcNameSty{Initialise} and {\MsetLeq}. In Algorithm
\ref{msoleq-initialize-entailed}, we show how we revise Algorithm
\ref{msoleq-initialize}. We add line 0 to initialise the flag
$entailed$ to $false$. We replace line 5 of Algorithm
\ref{msoleq-initialize} with lines 5-7. Before calling {\MsetLeq}
, we now initialise our new occurrence vectors
$occ(\myceiling(\myvec{X}))$ and $occ(\myfloor(\myvec{Y}))$ in a
similar way to that of $occ(\myfloor(\myvec{X}))$ and
$occ(\myceiling(\myvec{Y}))$: we create a pair of vectors
$\myvec{ex}$ and $\myvec{ey}$ of length $u-l+1$ where each $ex_i$
and $ey_i$ are first set to $0$. Then, for each value $v$ in
$\msetl \myceiling(\myvec{X}) \msetr$, we increment $ex_v$ by 1.
Similarly, for each $v$ in $\msetl \myfloor(\myvec{Y}) \msetr$, we
increment $ey_v$ by 1. These vectors are then used  in {\MsetLeq}
to detect entailment. It is possible to maintain $\myvec{ex}$ and
$\myvec{ey}$ incrementally. When the maximum value in some
$\domain{D}(X_i)$ changes, we update $\myvec{ex}$ by incrementing
the entry corresponding to new $max(X_i)$ by 1, and decrementing
the entry corresponding to old $max(X_i)$ by 1. Likewise, when the
minimum value in some $\domain{D}(Y_i)$ changes, we update
$\myvec{ey}$ by incrementing the entry corresponding to new
$min(Y_i)$ by 1, and decrementing the entry corresponding to old
$min(Y_i)$ by 1.

\begin{algorithm}[t!]
\begin{footnotesize}
 \SetLine
 \AlgData{$\la X_0, X_1, \ldots, X_{n-1} \ra$, $\la Y_0, Y_1, \ldots, Y_{n-1} \ra$}
 \AlgResult{GAC($\myvec{X} \leq_{m} \myvec{Y}$)}
 \numline{A0}{0}\lIf{$entailed$}{return\;}
 \numline{$\Rightarrow$}{0}\lIf{$\myvec{ex} \leq_{lex} \myvec{ey}$}{$entailed:=true$, return\;}
 \numline{A1}{0}\ProcNameSty{SetPointersAndFlags}\;
 \numline{B1}{0}\ForEach{$i \in [0,n)$}{
 \numline{B2}{1}\If{$min(X_i) \neq max(X_i)$}{
 \numline{B3}{2}\If{$min(X_i) \geq \alpha$}{
 \numline{$\Rightarrow$}{3}$ex_{max(X_i)}:=ex_{max(X_i)}-1$, \ProcNameSty{setMax($X_i,min(X_i)$)}\;
 \numline{$\Rightarrow$}{3}$ex_{max(X_i)}:=ex_{max(X_i)}+1$\;}
 \numline{B4}{2}\If{$max(X_i) \geq \alpha~\And~min(X_i)<\alpha$}{
 \numline{B5$\Rightarrow$}{3}$ex_{max(X_i)}:=ex_{max(X_i)}-1$, \ProcNameSty{setMax($X_i,\alpha$)}\;
                \vdots
 \numline{$\Rightarrow$}{3}$ex_{max(X_i)}:=ex_{max(X_i)}+1$\;
                }

} }
 \numline{$\Rightarrow$}{0}\lIf{$\myvec{ex} \leq_{lex} \myvec{ey}$}{$entailed:=true$, return\;}
 \numline{C1}{0}\ForEach{$i \in [0,n)$}{
 \numline{C2}{1}\If{$min(Y_i) \neq max(Y_i)$}{
 \numline{C3}{2}\If{$max(Y_i) >\alpha$}{
 \numline{$\Rightarrow$}{3}$ey_{min(Y_i)}:=ey_{min(Y_i)}-1$, \ProcNameSty{setMin($Y_i,max(Y_i)$)}, $ey_{min(Y_i)}:=ey_{min(Y_i)}+1$\;}
 \numline{C4}{2}\If{$max(Y_i)=\alpha~\And~min(Y_i) \leq \beta$}{
 \numline{C5}{3}\If{$ox_{\alpha}+1=oy_{\alpha}~\And~\gamma$}{
 \numline{C6$\Rightarrow$}{4}$ey_{min(Y_i)}:=ey_{min(Y_i)}-1$, \ProcNameSty{setMin($Y_i,\beta$)}\;
                \vdots
 \numline{$\Rightarrow$}{4}$ey_{min(Y_i)}:=ey_{min(Y_i)}+1$}
               }
 }
}
 \numline{$\Rightarrow$}{0}\lIf{$\myvec{ex}\leq_{lex}\myvec{ey}$}{$entailed:=true$, return\;}

\caption{\MsetLeq} \label{msoleq-propagate-entailed}
\end{footnotesize}
\end{algorithm}

In Algorithm \ref{msoleq-propagate-entailed}, we show how we
modify the filtering algorithm\index{Filtering algorithm} given in
Algorithm \ref{msoleq-propagate} to deal with the three possible
scenarios described above. We add line {\bf A0} where we return if
the constraint has already been entailed in the past. Moreover,
just before setting our pointers and flags, we check whether the
recent modifications that triggered the algorithm resulted in
entailment. If this is the case, we first set $entailed$ to $true$
and then return from the algorithm. Furthermore, we check
entailment after the algorithm goes through its variables. Lines
{\bf B1-B11} visit the variables of $\myvec{X}$ and prune
inconsistent values from the upper bounds, affecting $\myvec{ex}$.
Even if we have $\myvec{ex}>_{lex} \myvec{ey}$ when the algorithm
is called, we might get $\myvec{ex}\leq_{lex} \myvec{ey}$ just
before the algorithm proceeds to the variables of $\myvec{Y}$. In
such case, we return from the algorithm after setting $entailed$
to $true$. As an example, assume we have $\myvec{X} \leq_m
\myvec{Y}$, and {\MsetLeq} is called with $\myvec{X}=\la \{1,2\},
\{1,2,4\}\ra$ and $\myvec{Y}=\la \{2,3\}, \{2,3\}\ra$. As 4 in
$\domain{D}(X_1)$ lacks support, it is pruned. Now we have
$\myvec{ex}=\myvec{ey}$. Alternatively, the constraint might be
entailed after the algorithm visits the variables of $\myvec{Y}$
and prunes inconsistent values from the lower bounds, affecting
$\myvec{ey}$. In this case, we return from the algorithm by
setting $entailed$ to $true$. As an example, assume we also have
$0$ in $\domain{D}(Y_1)$ in the previous example. The constraint
is entailed only after the variables of $\myvec{Y}$ are visited
and $0$ is removed.

Finally, before/after the algorithm modifies $max(X_i)$ or
$min(Y_i)$ of some $i$ in $[0,n)$, we keep our occurrence vectors
$\myvec{ex}$ and $\myvec{ey}$ up-to-date by
decrementing/incrementing the necessary entries.

\section{Alternative Approaches}
\label{ch-gacmset:alternativeapproaches} There are several
alternative ways known for posting and propagating multiset ordering
constraints. We can, for instance, post arithmetic inequality constraints,
or decompose multiset ordering constraints
into other constraints. In this section, we
explore these approaches and argue why it is preferable
to propagate multiset ordering constraints using our filtering
algorithms.

\subsection{Arithmetic Constraint}
\label{ch-gacmset:alternativeapproaches-arth} We can achieve
multiset ordering between two vectors by assigning a weight to
each value, summing the weights along each vector, and then
insisting the sums to be non-decreasing. Since the ordering is
determined according to the maximum value in the vectors, the
weight should increase with the value. A suitable weighting scheme
was proposed in \cite{ks:symcon02}, where each value $v$ gets
assigned the weight $n^v$, where $n$ is the length of the vectors.
$\myvec{X} \leq_m \myvec{Y}$ on vectors of length $n$ can then be
enforced via the following arithmetic inequality
constraint\index{Multiset ordering!constraint!arithmetic}:
\[
n^{X_0} + \ldots +n^{X_{n-1}} \leq n^{Y_0} + \ldots +n^{Y_{n-1}}
\]
Therefore, a vector containing one element
with value $v$ and $n-1$ 0s is greater than a vector whose $n$
elements are only $v-1$. This is in fact similar to the
transformation of a leximin fuzzy CSP into an equivalent MAX CSP
\cite{schiex7}. Strict multiset ordering constraint $\myvec{X}<_m
\myvec{Y}$ is enforced by disallowing equality:
\[
n^{X_0} + \ldots +n^{X_{n-1}} < n^{Y_0} + \ldots +n^{Y_{n-1}}
\]
BC on such arithmetic constraints does the same pruning as GAC on
the original multiset ordering constraints. However, such
arithmetic constraints are feasible only for small $n$ and $u$,
where $u$ is the maximum value in the domains of the variables. As
$n$ and $u$ get large, $n^{X_i}$ or $n^{Y_i}$ will be a very large
number and therefore it might be impossible to implement the
multiset ordering constraint.
Consequently, it can be preferable to post and propagate the
multiset ordering constraints using our global constraints.

\begin{theorem} GAC(${\myvec{X} \leq_m \myvec{Y}}$) and GAC(${\myvec{X} <_m \myvec{Y}}$)
are equivalent to BC on the corresponding arithmetic constraints.
\end{theorem}
\proof We just consider GAC($\myvec X \leq_{m} \myvec Y)$ as the
proof for GAC($\myvec X <_{m} \myvec Y$) is entirely analogous. As
$\myvec{X} \leq_m \myvec{Y}$ and the corresponding arithmetic
constraint are logically equivalent, BC($\myvec X \leq_{m} \myvec
Y$) and BC on the arithmetic constraint are equivalent. By Theorem
\ref{th-BCmset}, BC($\myvec X \leq_{m} \myvec Y$) is equivalent to
GAC($\myvec X \leq_{m} \myvec Y$). \qed

\subsection{Decomposition}
\label{ch-gacmset:alternativeapproaches-dec} Global ordering
constraints can often be built out of the logical connectives
($\And$, $\Or$, $\Implies$, $\Iff$, and $\Not$) and  existing
(global) constraints. We can thus compose other constraints
between $\myvec{X}$ and $\myvec{Y}$ so as to obtain the multiset
ordering constraint between $\myvec{X}$ and $\myvec{Y}$. We refer
to such a logical constraint as a decomposition\index{Multiset
ordering!constraint!decomposition} of the multiset ordering
constraint.

The multiset view of two vectors of integers $\myvec{x}$ and
$\myvec{y}$ are multiset ordered $\msetl \myvec{x} \msetr \leq_m
\msetl \myvec{y} \msetr$ iff $occ(\myvec{x}) \leq_{lex}
occ(\myvec{y})$ by Theorem \ref{th-occ}. One way of decomposing
the multiset ordering constraint $\myvec{X} \leq_m \myvec{Y}$ is
thus insisting that the occurrence vectors associated with the
vectors assigned to $\myvec{X}$ and $\myvec{Y}$ are
lexicographically ordered. Such occurrence vectors can be
constructed via an extended global cardinality constraint
($gcc$)\index{Global cardinality constraint}\index{gcc@$gcc$|see
{Global cardinality constraint}}. Given a vector of variables
$\myvec{X}$ and a vector of values $\myvec{d}$, the constraint
$gcc(\myvec{X},\myvec{d},\myvec{OX})$ ensures that $OX_i$ is the
number of variables in $\myvec{X}$ assigned to $d_i$. To ensure
multiset ordering, we can enforce lexicographic ordering
constraint on a pair of occurrence vectors constructed via $gcc$
where $\myvec{d}$ is the vector of values that the variables can
be assigned to, arranged in descending order, without any
repetition:
\[
gcc(\myvec{X},\myvec{d},\myvec{OX})~\And~gcc(\myvec{Y},\myvec{d},\myvec{OY})~\And~\myvec{OX}
\leq_{lex} \myvec{OY} \]
In order to decompose the strict multiset
ordering constraint $\myvec{X} <_m \myvec{Y}$, we need to enforce
strict lexicographic ordering constraint on the occurrence
vectors:
\[
gcc(\myvec{X},\myvec{d},\myvec{OX})~\And~gcc(\myvec{Y},\myvec{d},\myvec{OY})~\And~\myvec{OX}
<_{lex} \myvec{OY}
\]
We call this way of decomposing a multiset ordering constraint as
$gcc$ decomposition.

The $gcc$ constraint is available in, for instance, ILOG Solver 5.3
\cite{IlogSolver}, SICStus Prolog 3.10.1 \cite{SicstusProlog}, and
the FaCiLe constraint solver 1.0 \cite{facile}. These solvers
propagate the $gcc$ constraint using the algorithm proposed in
\cite{regin2}. Among the various filtering algorithms of $gcc$,
which maintain either GAC \cite{regin2}\cite{gcc:bc1} or BC
\cite{gcc:bc1}\cite{gcc:bc3}, only the algorithms in \cite{gcc:bc3}
prune values from $\myvec{OX}$ and $\myvec{OY}$. Even though the
algorithm integrated in ILOG Solver 5.3 may also prune the
occurrence vectors, this may not always be the case. For instance,
when we have $gcc(\la
\{1\},\{1,2\},\{1,2\},\{2\},\{3,4\},\{3,4\}\ra, \la 4,3,2,1\ra, \la
\{1\},\{1\},\{1,2\},\\\{1,2,3\}\ra$, ILOG Solver 5.3 leaves
$\myvec{OX}$ unchanged even though $1$ in $\domain{D}(OX_3)$ is not
consistent. This shows that there is currently very limited support
in the constraint toolkits to propagate the multiset ordering
constraint using the $gcc$ decomposition. Also, as the following
theorems demonstrate, the $gcc$ decomposition of a multiset ordering
constraint hinders constraint propagation.

\begin{theorem}
\label{th-dec1} GAC($\myvec{X} \leq_m \myvec{Y})$ is strictly
stronger than
GAC(gcc($\myvec{X},\myvec{d},\myvec{OX}$)), GAC(gcc($\myvec{Y},\\
\myvec{d},\myvec{OY}$)), and GAC($\myvec{OX} \leq_{lex}
\myvec{OY}$), where $\myvec{d}$ is the vector of values that the
variables can take, arranged in descending order, without any
repetition.
\end{theorem}
\proof Since $\myvec{X} \leq_{m} \myvec{Y}$ is GAC, every value
has a support $\myvec{x}$ and $\myvec{y}$ where $occ(\myvec{x})
\leq_{lex} occ(\myvec{y})$, in which case all the three
constraints posted in the decomposition are satisfied. Hence,
every constraint imposed is GAC, and GAC($\myvec{X} \leq_m
\myvec{Y})$ is as strong as its decomposition. To show strictness,
consider $\myvec{X}=\la \{0,3\},\{2\} \ra$ and $\myvec{Y}=\la
\{2,3\},\{1\} \ra$. The multiset ordering constraint $\myvec{X}
\leq_m \myvec{Y}$ is not GAC as $3$ in $\domain{D}(X_0)$ has no
support. By enforcing GAC($gcc$($\myvec{X},\la
3,2,1,0\ra,\myvec{OX}$)) and GAC($gcc$($\myvec{Y},\la
3,2,1,0\ra,\myvec{OY}$)) we obtain the following occurrence
vectors:
\[
\begin{array}{cccccccccccccccccccc}
 \myvec{OX} & = & \la \{0,1\}, & \{1\}, & \{ 0\}, &\{0,1\} \ra \\
 \myvec{OY} & = & \la \{0,1\}, & \{0,1\}, & \{1\}, &\{0\} \ra
\end{array}
\]
Since we have GAC($\myvec{OX} \leq_{lex} \myvec{OY}$), $\myvec{X}$
and $\myvec{Y}$ remain unchanged. \qed

\begin{theorem}
GAC($\myvec{X} <_m \myvec{Y})$ is strictly stronger than
GAC(gcc($\myvec{X},\myvec{d},\myvec{OX}$)), GAC(gcc($\myvec{Y},\\
\myvec{d},\myvec{OY}$)), and GAC($\myvec{OX} <_{lex} \myvec{OY}$),
where $\myvec{d}$ is the vector of values that the variables can
take, arranged in descending order, without any repetition.
\end{theorem}
\proof The example in Theorem \ref{th-dec1} shows the strictness.
\qed

In Theorem \ref{th-sort}, we have established that $occ(\myvec{x})
\leq_{lex} occ(\myvec{y})$ iff $sort(\myvec{x}) \leq_{lex}
sort(\myvec{y})$. Putting Theorems \ref{th-occ} and \ref{th-sort}
together, the multiset view of two vectors of integers $\myvec{x}$
and $\myvec{y}$ are multiset ordered $\msetl \myvec{x} \msetr
\leq_m \msetl \myvec{y} \msetr$ iff $sort(\myvec{x}) \leq_{lex}
sort(\myvec{y})$. This suggests another way of decomposing a
multiset ordering constraint $\myvec{X} \leq_m \myvec{Y}$: we
insist that the sorted versions of the vectors assigned to
$\myvec{X}$ and $\myvec{Y}$ are lexicographically ordered. For
this purpose, we can use the constraint
$sorted$\index{sorted@$sorted$ constraint} which is available in,
for instance, ECLiPSe constraint solver 5.6 \cite{Eclipse},
SICStus Prolog 3.10.1 \cite{SicstusProlog}, and
the FaCiLe constraint solver 1.0 \cite{facile}. Given a vector of
variables $\myvec{X}$, $sorted(\myvec{X},\myvec{SX})$ ensures that
$\myvec{SX}$ is of length $n$ and is a sorted permutation of
$\myvec{X}$. To ensure multiset ordering, we can enforce
lexicographic ordering constraint on a pair of vectors which are
constrained to be the sorted versions of the original vectors in
descending order:
\[
sorted(\myvec{X},\myvec{SX})~\And~sorted(\myvec{Y},\myvec{SY})~\And~\myvec{SX}
\leq_{lex} \myvec{SY}
\]
A strict multiset ordering constraint $\myvec{X} <_m \myvec{Y}$ is
then achieved by enforcing strict lexicographic ordering
constraint on the sorted vectors:
\[
sorted(\myvec{X},\myvec{SX})~\And~sorted(\myvec{Y},\myvec{SY})~\And~\myvec{SX}
<_{lex} \myvec{SY}
\]
We call this way of decomposing a multiset ordering constraint as
the $sort$ decomposition.

The $sorted$ constraint has previously been studied and some BC
filtering algorithms have been proposed
\cite{sorted2}\cite{sorted1}\cite{sorted3}. Unfortunately, we lose
in the amount of constraint propagation also by the $sort$
decomposition of a multiset ordering constraint.

\begin{theorem}
\label{th-dec2} GAC($\myvec{X} \leq_m \myvec{Y})$ is strictly
stronger than GAC(sorted($\myvec{X},\myvec{SX}$)),
GAC(sorted\\($\myvec{Y},\myvec{SY}$)), and GAC($\myvec{SX}
\leq_{lex} \myvec{SY}$).
\end{theorem}
\proof Since $\myvec{X} \leq_{m} \myvec{Y}$ is GAC, every value
has a support $\myvec{x}$ and $\myvec{y}$ where $sort(\myvec{x})
\leq_{lex} sort(\myvec{y})$, in which case all the three
constraints posted in the decomposition are satisfied. Hence,
every constraint imposed is GAC, and GAC($\myvec{X} \leq_m
\myvec{Y})$ is as strong as its decomposition. To show strictness,
consider $\myvec{X}=\la \{0,3\},\{2\} \ra$ and $\myvec{Y}=\la
\{2,3\},\{1\} \ra$. The multiset ordering constraint $\myvec{X}
\leq_m \myvec{Y}$ is not GAC as $3$ in $\domain{D}(X_0)$ has no
support. By enforcing GAC($sorted$($\myvec{X},\myvec{SX}$)) and
GAC($sorted$($\myvec{Y},\myvec{SY}$)) we obtain the following
vectors:
\[
\begin{array}{cccccccccccccccccccc}
 \myvec{SX} & = & \la \{2,3\}, & \{0,2\}\ra \\
 \myvec{SY} & = & \la \{2,3\}, & \{1\}  \ra
\end{array}
\]
Since we have GAC($\myvec{SX} \leq_{lex} \myvec{SY}$), $\myvec{X}$
and $\myvec{Y}$ remain unchanged. \qed

\begin{theorem}
GAC($\myvec{X} <_m \myvec{Y})$ is strictly stronger than
GAC(sorted($\myvec{X},\myvec{SX}$)),
GAC(sorted\\($\myvec{Y},\myvec{SY}$)), and GAC($\myvec{SX} <_{lex}
\myvec{SY}$).
\end{theorem}
\proof The example in Theorem \ref{th-dec2} shows strictness. \qed

How do the two decompositions compare? Assuming that GAC is
enforced on every $n$-ary constraint of a decomposition, the
$sort$ decomposition is superior to the $gcc$ decomposition.

\begin{theorem}
\label{sort-gcc} The $sort$ decomposition of $\myvec{X} \leq_m
\myvec{Y}$ is strictly stronger than the $gcc$ decomposition of
$\myvec{X} \leq_m \myvec{Y}$.
\end{theorem}
\proof Assume that a value is pruned from $\myvec{X}$ due to the
$gcc$ decomposition. Then, there is an index $\alpha$ such that
$\Not(OX_\alpha \groundandequal OY_\alpha)$ and for all $i>\alpha$
we have $OX_i \groundandequal OY_i$. Moreover, we have
$min(OX_i)=max(OY_i)$ and $max(OX_i)>max(OY_i)$. The reason is
that, only in this case, GAC($\myvec{OX} \leq_{lex} \myvec{OY}$)
will not only prune values from $OX_\alpha$ but also from
$\myvec{X}$. In any other case, we will either get no pruning at
$OX_\alpha$, or the pruning at $OX_\alpha$ will reduce the number
of occurrences of $\alpha$ in $\myvec{X}$ without deleting any of
$\alpha$ from $\myvec{X}$. Now consider the vectors $\myvec{SX}$
and $\myvec{SY}$. We name the index of $\myvec{SX}$ and
$\myvec{SY}$, where $\alpha$ first appears in the domains of
$\myvec{SX}$ and $\myvec{SY}$, as $i$. Since the number of
occurrences of any value greater than $\alpha$ is already
determined and is the same in both $\myvec{X}$ and $\myvec{Y}$,
the subvectors of $\myvec{SX}$ and $\myvec{SY}$ above $i$ are
ground and equal. For all $i \leq j < i+min(OX_i)$, we have $SX_j
\groundandequal SY_j \assigned \alpha$. Since
$max(OX_i)>max(OY_i)$, at position $k=i+min(OX_i)$ we will have
$\alpha$ in $\domain{D}(SX_k)$ but not in $\domain{D}(SY_k)$ whose
values are less than $\alpha$. To have $\myvec{SX} \leq_{lex}
\myvec{SY}$, $\alpha$ in $\domain{D}(SX_k)$ is eliminated. This
propagates to the pruning of $\alpha$ from the remaining variables
of $\myvec{SX}$, as well as from domains of the uninstantiated
variables of $\myvec{X}$. Hence, any value removed from
$\myvec{X}$ due to the $gcc$ decomposition is removed from
$\myvec{X}$ also by the $sort$ decomposition. The proof can easily
be reverted for values being removed from $\myvec{Y}$.

To show that the $sort$ decomposition dominates the $gcc$
decomposition, consider $\myvec{X}=\la \{1,2\} \ra$ and
$\myvec{Y}=\la \{0,1,2\} \ra$ where $0$ in $\domain{D}(Y_0)$ is
inconsistent and therefore $\myvec{X} \leq_m \myvec{Y}$ is not
GAC. We have $\myvec{SX}=\la \{1,2\} \ra$ and $\myvec{SY}=\la
\{0,1,2\} \ra$ by GAC($sorted$($\myvec{X},\myvec{SX}$)) and
GAC($sorted$($\myvec{Y},\myvec{SY}$)), and $\myvec{OX}=\la
\{0,1\}, \{0,1\},\{0\} \ra$ and $\myvec{OY}=\la \{0,1\}, \{0,1\},
\{0,1\} \ra$ by GAC($gcc$($\myvec{X},\la 2, 1, 0 \ra,\myvec{OX}$))
and GAC($gcc$($\myvec{Y},\la 2, 1, 0 \ra, \myvec{OY}$)). To
achieve GAC($\myvec{SX} \leq_{lex} \myvec{SY}$), $0$ in
$\domain{D}(SY_0)$ is pruned. This leads to the pruning of $0$
also from $\domain{D}(Y_0)$ so as to establish
GAC($sorted$($\myvec{Y},\myvec{SY}$)). On the other hand, we have
GAC($\myvec{OX} \leq_{lex} \myvec{OY}$), in which case no value is
pruned from any variable. \qed

\begin{theorem}
The $sort$ decomposition of $\myvec{X} <_m \myvec{Y}$ is strictly
stronger than the $gcc$ decomposition of $\myvec{X} <_m
\myvec{Y}$.
\end{theorem}
\proof The example in Theorem \ref{sort-gcc} shows strictness.
\qed

Even though the $sort$ decomposition of $\myvec{X} \leq_m
\myvec{Y}$ is stronger than the $gcc$ decomposition of $\myvec{X}
\leq_m \myvec{Y}$, GAC on $\myvec{X} \leq_m \myvec{Y}$ can lead to
more pruning than any of the two decompositions. A similar
argument holds also for $\myvec{X} <_m \myvec{Y}$. Hence, it can
be preferable to post and propagate multiset ordering constraints
via our global constraints.

\section{Multiple Vectors}
\label{ch-gacmset:multiple vectors}

We often have multiple multiset ordering constraints. For example,
we post multiset ordering constraints on the rows or columns of a
matrix of decision variables because we want to break row or
column symmetry. We can treat such a problem as a single global
ordering constraint over the whole matrix. Alternatively, we can
decompose it into multiset ordering constraints between adjacent
or all pairs of vectors. In this section, we demonstrate that such
decompositions hinder constraint propagation.

The following theorems hold for $n$ vectors of $m$ constrained
variables.

\begin{theorem}
\label{th-mset-GACcons1} GAC($\myvec X_i \leq_{m} \myvec X_j$) for
all $0\leq i<j\leq n-1$ is strictly stronger than GAC($\myvec X_i
\leq_{m} \myvec X_{i+1}$) for all $0 \leq i <n-1$.
\end{theorem} \proof GAC($\myvec X_i \leq_{m} \myvec X_j$) for
all $0\leq i<j\leq n-1$ is as strong as GAC($\myvec X_i \leq_{m}
\myvec X_{i+1}$) for all $0 \leq i <n-1$, because the former
implies the latter. To show strictness, consider the following $3$
vectors:
\[
\begin{array}{ccccccccccccccc}
  \myvec X_0 & = & \langle\{0,3\},&\{2\} \rangle \\
  \myvec X_1 & = & \langle\{0,1,2,3\},&\{0,1,2,3\} \rangle \\
  \myvec X_2 & = & \langle\{2,3\},&\{1\} \rangle
\end{array}
\]
We have GAC($\myvec X_i \leq_{m} \myvec X_{i+1}$) for all $0 \leq
i <2$. The assignment $X_{0,0} \assigned 3$ forces $\myvec{X_0}$
to be $\la 3, 2\ra$, and we have $\myceiling(\myvec{X_2})=\la 3,1
\ra$. Since $\msetl 3 ,2 \msetr >_m \msetl 3,1 \msetr$,
GAC($\myvec X_0 \leq_{m} \myvec X_2$) does not hold. \qed

\begin{theorem}
\label{th-mset-GACcons2} GAC($\myvec X_i <_{m} \myvec X_j$) for
all $0\leq i<j\leq n-1$ is strictly stronger than GAC($\myvec X_i
<_{m} \myvec X_{i+1}$) for all $0 \leq i <n-1$.
\end{theorem}
\proof The example in Theorem \ref{th-mset-GACcons1} shows
strictness. \qed

\begin{theorem}
\label{th-mset-GACevery1} GAC($\forall ij~0\leq i<j\leq n-1
\Dot~\myvec X_i\leq_{m} \myvec X_j$) is strictly stronger than
GAC($\myvec X_i \leq_{m} \myvec X_j$) for all $0\leq i<j\leq n-1$.
\end{theorem}
\proof GAC($\forall ij~0\leq i<j\leq n-1 \Dot~\myvec X_i\leq_{m}
\myvec X_j$) is as strong as GAC($\myvec X_i \leq_{m} \myvec X_j$)
for all $0\leq i<j\leq n-1$, because the former implies the
latter. To show strictness, consider the following $3$ vectors:
\[
\begin{array}{cccccccccccc}
  \myvec X_0 & = & \langle\{0,3\},&\{1\} \rangle \\
  \myvec X_1 & = & \langle\{0,2\},&\{0,1,2,3\} \rangle \\
  \myvec X_2 & = & \langle\{0,1\},&\{0,1,2,3\}  \rangle
  \end{array}
\]
We have  GAC($\myvec X_i \leq_{m} \myvec X_j$) for all $0\leq
i<j\leq 2$. The assignment $X_{0,0} \assigned 3$ is supported by
$X_0 \assigned \la 3,1 \ra$, $X_1 \assigned \la 2,3 \ra$, and $X_2
\assigned \la 1,3 \ra$. In this case, $\myvec{X_1} \leq_m
\myvec{X_2}$ is $false$. Therefore, GAC($\forall ij~0\leq i<j\leq
2 \Dot~\myvec X_i\leq_{m} \myvec X_j$) does not hold. \qed

\begin{theorem}
\label{th-mset-GACevery2} GAC($\forall ij~0\leq i<j\leq n-1
\Dot~\myvec X_i<_{m} \myvec X_j$) is strictly stronger than
GAC($\myvec X_i <_{m} \myvec X_j$) for all $0\leq i<j\leq n-1$.
\end{theorem}
\proof GAC($\forall ij~0\leq i<j\leq n-1 \Dot~\myvec X_i<_{m}
\myvec X_j$) is as strong as GAC($\myvec X_i <_{m} \myvec X_j$)
for all $0\leq i<j\leq n-1$, because the former implies the
latter. To show strictness, consider the following $3$ vectors:
\[
\begin{array}{cccccccccccc}
  \myvec X_0 & = & \langle\{0,3\},&\{1\} \rangle \\
  \myvec X_1 & = & \langle\{1,3\},&\{0,1,3\} \rangle \\
  \myvec X_2 & = & \langle\{0,2\},&\{0,1,2,3\}  \rangle
  \end{array}
\]
We have  GAC($\myvec X_i <_{m} \myvec X_j$) for all $0\leq i<j\leq
2$. The assignment $X_{0,0} \assigned 3$ is supported by $X_0
\assigned \la 3,1 \ra$, $X_1 \assigned \la 3,3 \ra$, and $X_2
\assigned \la 2,3 \ra$. In this case, $\myvec{X_1} <_m
\myvec{X_2}$ is $false$. Therefore, GAC($\forall ij~0\leq i<j\leq
2 \Dot~\myvec X_i<_{m} \myvec X_j$) does not hold. \qed

\section{Experiments}
\label{ch-gacmset:experiments} We implemented our global constraints
$\leq_m$ and $<_m$ in C++ using ILOG Solver 5.3 \cite{IlogSolver}.
Due the absence of the $sorted$ constraint in Solver 5.3, the
multiset ordering constraint is decomposed via the $gcc$
decomposition using the {\bf IloDistribute} constraint. This
constraint is the $gcc$ constraint but it does not always prune
completely the occurrence vectors as described before.

In the experiments, we have a matrix of decision variables where the
rows and/or columns are (partially) symmetric. To break the
symmetry, we post multiset ordering constraints on the adjacent
symmetric rows or columns, and address several questions in the
context of looking for one
solution or the optimal solution. First, does our filtering
algorithm(s) do more inference in practice than its decomposition?
Similarly, is the algorithm more efficient in practice than its
decomposition? Second, is it feasible to post the arithmetic
constraint? How does our algorithm compare to BC on the arithmetic
constraint? Even though studying the effectiveness of the multiset
ordering constraints in breaking symmetry is out of the scope of
this paper, we provide experimental evidence of their value
in symmetry breaking.

We report experiments on three problem domains: the progressive
party problem,  the rack configuration problem, and the sport
scheduling problem. The decisions made when modelling and solving a
problem are tuned by our initial experimentation. The results are
shown in tables where a ``-'' means no result is obtained in 1 hour
(3600 secs). The best result of each entry in a table is typeset in
bold. If posing an ordering constraint on the rows (resp. columns)
is done via a technique called $Tech$ then we write $Tech$ R (resp.
$Tech$ C). The ordering constraints are enforced just between the
adjacent rows and/or columns as we have found it not worthwhile to
post them between  all pairs.

Finally, the hardware used for the experiments is a 1Ghz pentium III
processor with 256Mb RAM running Windows XP.

\subsection{Progressive Party Problem} The {\bf progressive party
problem} arises in the context of organising the social programme
for a yachting rally (prob013 in CSPLib). We consider a variant of
the problem proposed in  \cite{sbhw96}.
There is a set $\mydomain{Hosts}$
of host boats 
and a set $\mydomain{Guests}$ of guest boats. 
Each host boat $i$ is characterised by a tuple $\la hc_i, c_i\ra$,
where and $hc_i$ is its crew size and $c_i$ is its capacity; and
each guest boat is described by $gc_i$ giving its crew size. The
problem is to assign hosts to guests over $p$ time periods such
that:
\begin{itemize}
\item a guest crew never visits the same host twice;
\item no two guest crews meet more than once;
\item the spare capacity of each host boat, after accommodating its
own crew, is not exceeded.
\end{itemize}

\begin{figure}[t!]
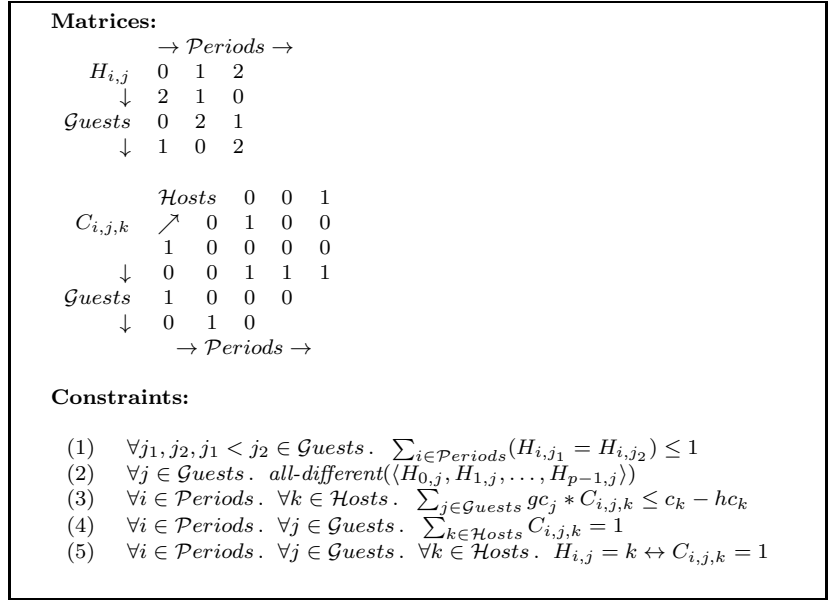

\begin{center}
\framebox{ \begin{scriptsize}
\begin{tabular}{l}
\mbox{{\bf Matrices:}} \\
$
\begin{array}{rcccc}
& \multicolumn{4}{c}{\mbox{$\rightarrow \mydomain{Periods} \rightarrow$ }} \\
H_{i,j}      &0&1&2\\
\downarrow   &2&1&0\\
\mydomain{Guests}&0&2&1\\
\downarrow   &1&0&2
\end{array}$\\\\
$
\begin{array}{rccccc}
& \multicolumn{2}{c}{\mydomain{Hosts}}&0&0&1\\
C_{i,j,k}& \nearrow   &0&1&0&0\\
                &1&0&0&0&0\\
\downarrow           &0&0&1&1&1\\
\mydomain{Guests}        &1&0&0&0\\
\downarrow           &0&1&0\\
& \multicolumn{5}{c}{\mbox{$\rightarrow \mydomain{Periods} \rightarrow$}}  \\
\end{array}$\\\\
\mbox{\bf Constraints:} \\\\
$
\begin{tabular}{ll}
(1)     & $\forall j_1, j_2, j_1 <j_2 \in \mydomain{Guests} \Dot~\sum_{i \in \mydomain{Periods}} (H_{i, j_1}=H_{i, j_2}) \leq 1$ \\
(2)     & $\forall j \in \mydomain{Guests} \Dot~ \myalldiff (\la H_{0,j}, H_{1,j}, \ldots,  H_{p-1,j} \ra)$\\
(3)     & $\forall i \in \mydomain{Periods} \Dot~\forall k \in \mydomain{Hosts} \Dot~\sum_{j \in \mydomain{Guests}} {gc}_j * C_{i,j,k} \leq c_k -hc_k$\\
(4)     & $\forall i \in \mydomain{Periods} \Dot~\forall j \in \mydomain{Guests} \Dot~\sum_{k \in \mydomain{Hosts}} C_{i,j,k} = 1$\\
(5)     & $\forall i \in \mydomain{Periods} \Dot~\forall j \in \mydomain{Guests} \Dot~\forall k \in \mydomain{Hosts} \Dot~H_{i,j}=k \Iff C_{i,j,k}=1$ \\
\end{tabular}$ \\\\
\end{tabular}
\end{scriptsize}
}
\end{center}
\caption{The matrix model of the progressive party problem in
\cite{sbhw96}.} \label{ch-matrixmodels:table-ppp}
\end{figure}

A matrix model of this problem is given in \cite{sbhw96}. It has a
2-d matrix $H$ to represent the assignment of hosts to  guests in
time periods (see Figure \ref{ch-matrixmodels:table-ppp}). The
matrix $H$ is indexed by the set $\mydomain{Periods}$ of time
periods and $\mydomain{Guests}$, taking values from
$\mydomain{Hosts}$. The first constraint enforces that two guests
can meet at most once by introducing a new set of 0/1 variables:

$$\forall i \in \mydomain{Periods} \Dot~\forall j_1, j_2, j_1<j_2 \in \mydomain{Guests} \Dot ~ M_{i,j1,j2}=1 \Iff H_{i,j1}=H_{i,j2}$$
The sum of these new variables are then constrained to be at most 1.
The $\myalldiff$ constraints on the rows of this matrix ensure that
no guest revisits a host. Additionally, a 3-d 0/1 matrix $C$ of
$\mydomain{Periods} \times \mydomain{Guests} \times
\mydomain{Hosts}$ is used. A variable $C_{i,j,k}$ in this new matrix
is $1$ iff the host boat $k$ is visited by guest $j$ in period $i$.
Even though $C$ replicates the information held in the 2-d matrix,
it allows capacity constraints to be stated concisely. The sum
constraints on $C$ ensure that a guest is assigned to exactly one
host on a time period. Finally, channelling constraints are used to
link the variables of $H$ and $C$.

The time periods as well as the guests with equal crew size are
indistinguishable. Hence, this model of the problem has partial row
symmetry between the indistinguishable guests of $H$, and column
symmetry.
In the following we first show that multiset ordering constraints
are useful in breaking index symmetry.

To break the row and column symmetries, we can utilise both
lexicographic ordering and multiset ordering constraints, as well as
combine lexicographic ordering constraints in one dimension of the
matrix with multiset ordering constraints in the other. Due to the
problem constraints, no pair of rows/columns can have equal
assignments, but they can be equal when viewed as multisets. This
gives us the models $<_{lex}$RC, $\leq_{m}$RC, $\leq_{m}$R
$\geq_{m}$C, $\leq_{m}$R $<_{lex}$C, $\leq_{m}$R $>_{lex}$C,
$<_{lex}$R $\leq_{m}$C, and $<_{lex}$R $\geq_{m}$C. As the matrix
$H$ has partial row symmetry, the ordering constraints on the rows
are posted on only the symmetric rows. The ordering constraints on
the columns are, however, posted on all the columns.

In our experiments, we compare the models described above in
contrast to the initial model of the problem in which no symmetry
breaking  ordering constraints are imposed. We consider the original
instance of the progressive party problem described in
\cite{sbhw96}, with 5 and 6 time periods.
As in \cite{sbhw96}, we give priority to the largest
crews, so the guest boats are ordered in descending order of their
size. Also, when assigning a host to a guest, we try a value first
which is most likely to succeed. We therefore order the host boats
in descending order of their spare capacity. We adopt two static
variable orderings, and instantiate $H$ either along its rows from
top to bottom, or along its columns from left to right.

\begin{table}[t]
\begin{footnotesize}
\begin{center}
\begin{tabular}{||l||r|r|l||r|r|l||}
\hline & \multicolumn{6} {|c||} {Problem} \\
        Model &\multicolumn{3}{|c||}{$5$ Time Periods}& \multicolumn{3}{|c||}{$6$ Time Periods}\\
       &Fails&Choice &Time &Fails&Choice & Time \\
       &     &points &(secs.)&    &points & (secs.) \\
\hline No symmetry breaking &180,738& 180,860& 75.9 & - & - & - \\
\hline $<_{lex}$RC &2,720 & 2,842& 2.7 & - & - & -\\
\hline $\leq_{m}$RC &- & -& - & - & - & -\\
\hline $\leq_{m}$R $\geq_{m}$C &9,207 & 9,329 & 8.0 & - & - & -\\

\hline $\leq_{m}$R $<_{lex}$C  &10,853 & 10,977& 8.6 & - & - & -\\
\hline $\leq_{m}$R $>_{lex}$C  &2,289 & 2,405& 2.6 & - & - & -\\

\hline $<_{lex}$R $\leq_{m}$C   &{\bf 2,016} & {\bf 2,137}& {\bf 2.0} & - & - & -\\
\hline $<_{lex}$R $\geq_{m}$C   &- & -& - & - & - & -\\

\hline
\end{tabular}
\end{center}
\caption{\label{ch-effectiveness:table:PPPResults1} Progressive
party problem with row-wise labelling of $H$.} \end{footnotesize}

\end{table}

\begin{table}[t!]
\begin{footnotesize}
\begin{center}
\begin{tabular}{||l||r|r|l||r|r|l||}

\hline & \multicolumn{6} {|c||} {Problem} \\
        Model &\multicolumn{3}{|c||}{$5$ Time Periods}& \multicolumn{3}{|c||}{$6$ Time Periods}\\
       &Fails&Choice &Time &Fails&Choice & Time \\
       &     &points &(secs.)&    &points & (secs.) \\
\hline No symmetry breaking &20,546& 20,676 & 9.0 & 20,722& 20,871&12.3 \\
\hline $<_{lex}$RC &20,546 & 20,676& 9.0 &20,722& 20,871&12.4 \\
\hline $\leq_{m}$RC &- & -& -& -& -&- \\
\hline $\leq_{m}$R $\geq_{m}$C &- & -& -& -& -&- \\

\hline $\leq_{m}$R $<_{lex}$C  &- & -& -& -& -&- \\
\hline $\leq_{m}$R $>_{lex}$C  &- & -& -& -& -&- \\

\hline $<_{lex}$R $\leq_{m}$C   &{\bf 7,038} & {\bf 7,168}& {\bf 3.4}& {\bf 7,053} & {\bf 7,202} & {\bf 4.6} \\
\hline $<_{lex}$R $\geq_{m}$C   &- & -& -& - & - & - \\

\hline
\end{tabular}
\end{center}
\caption{\label{ch-effectiveness:table:PPPResults2} Progressive
party problem with column-wise labelling of $H$.} \end{footnotesize}
\end{table}

The results of the experiments are shown in Tables
\ref{ch-effectiveness:table:PPPResults1} and
\ref{ch-effectiveness:table:PPPResults2}. With row-wise labelling of
$H$, we cannot solve the problem with 6 time periods with or without
the symmetry breaking ordering constraints. As for the other
instance, whilst many of the models we have considered give
significantly smaller search trees and shorter run-times,
$\leq_{m}$RC and $<_{lex}$R $\geq_{m}$C cannot return an answer
within an hour time limit. The smallest search tree and also the
shortest solving time is obtained by $<_{lex}$R $\leq_{m}$C, in
which case the reduction in the search effort is noteworthy compared
to the model in which no ordering constrains are imposed. This
supports our conjecture that lexicographic ordering constraints in
one dimension of a matrix combined with multiset ordering
constraints in the other can break more symmetry than lexicographic
ordering or multiset ordering constraints on both dimensions.


Next, we show that our filtering algorithm is the best way to
propagate multiset ordering constraints. To simplify the
presentation, we address only the row symmetry.
Given a set of indistinguishable guests $\{g_i, g_{i+1}, \ldots, g_j
\}$, we insist that  the rows corresponding to such guests are
multiset ordered: $\myvec{R_i} \leq_{m} \myvec{R_{i+1}} \ldots
\leq_{m} \myvec{R_j}$.
We impose such constraints by either using our filtering algorithm
{\MsetLeq}, or the $gcc$ decomposition, or the arithmetic
constraint.

We now consider several instances of the problem using the problem
data given in CSPLib. We randomly select the host boats in such a
way that the total spare capacity of the host boats is sufficient to
accommodate all the guests.
Table \ref{ch:gacmset:datappp} shows the data. The last column of
Table \ref{ch:gacmset:datappp} gives the percentage of the total
capacity used, which is a measure of constrainedness
\cite{walser:99}. We instantiate $H$ row-wise following the same
protocol described previously.

\begin{table}[t]
\begin{footnotesize}
\begin{center}
\begin{tabular}{||c|l|c|c|c||}
\hline Instance & Host Boats & Total Host & Total Guest& $\%$Capacity\\
        $\#$       &            & Spare Capacity   & Size       &            \\
\hline
1 & 2-12, 14, 16 & 102 & 92 & .90 \\
2 & 3-14, 16    & 100  & 90 & .90 \\
3 & 3-12, 14, 15, 16 & 101 & 91 & .90 \\
\hline
4 & 3-12, 14, 16, 25 & 101 & 92 & .91 \\
5 & 3-12, 14, 16, 23 & 99  & 90 & .91 \\
6 & 3-12, 15, 16, 25 & 100 & 91 & .91 \\
\hline
7 & 1, 3-12, 14, 16 & 100 & 92 & .92 \\
8 & 3-12, 16, 25, 26 & 100 & 92 & .92 \\
9 & 3-12, 14, 16, 30 & 98 & 90 & .92 \\
\hline
\end{tabular}
\end{center}
\caption{\label{ch:gacmset:datappp} Instance specification for the
progressive party problem.}
\end{footnotesize}
\end{table}



The results of the experiments are shown in Table
\ref{ch:gacmset:PPPResults2}. Note that all the problem instances
are solved for 5 time periods.
%
The results show that {\MsetLeq} maintains a significant advantage
over the $gcc$ decomposition and the arithmetic constraint. The
solutions to the instances, which can be solved within an hour
limit, are found quicker and compared to the $gcc$ decomposition
with much less failures. Note that {\MsetLeq} and the arithmetic
constraint methods create the same search tree.


\begin{table}[t]
\begin{footnotesize}
\begin{center}
\begin{tabular}{||c||ccc|c|ccc||}
\hline  & \multicolumn{3}{|c|}{{\MsetLeq} R}& \multicolumn{1}{|c|}{Arithmetic} & \multicolumn{3}{|c||}{$gcc$ R}\\
        Instance      &         &&                          &                     {Constraint R}& &&\\
        $\#$ & Fails & Choice & Time & Time & Fails & Choice & Time  \\
        &       & points & (secs.)& (secs.) &     & points & (secs.) \\
\hline 1 & {\bf 10,839 }& {\bf 10,963} & {\bf 8.3} & 16 & 20,367 & 20,491 &  11.6  \\
2 & {\bf 56,209}& {\bf 56,327} & {\bf 46.8} & 123.7 &  57,949 & 58,067 & 48.6 \\
3 & {\bf 27,461} & {\bf 27,575} & {\bf 17.1} & 39.1 & 42,741 & 42,855 & 20.5 \\
\hline 4 & {\bf 420,774} & {\bf 420,888} & {\bf 280.5} & 621.7 & 586,902 & 587,016 & 298.1 \\
5 & - & - & - & - & - & - & -\\
6 & {\bf 5,052} & {\bf 5,170} & {\bf 3.8} & 7.3 & 8,002 & 8,123 & 4.3 \\
\hline 7 & {\bf 86,432} & {\bf 86,547} & {\bf 65.5} & 135.2 & 128,080 & 128,195 & 75.7\\
8 & -& -& -& -& - & - &- \\
9 & -& -& -& -& - & - &- \\
\hline
\end{tabular}
\end{center}
\caption{\label{ch:gacmset:PPPResults2} Progressive party problem:
{\MsetLeq} vs $gcc$ decomposition and the arithmetic constraint with
row-wise labelling.}
\end{footnotesize}
\end{table}

\subsection{Rack Configuration Problem} The {\bf rack configuration
problem} consists of plugging a set of electronic cards into racks
with electronic connectors (prob031 in CSPLib). Each card is a
certain card type. A card type $i$ in the set $\mydomain{Ctypes}$
is characterised by a tuple $\la cp_i, d_i\ra$, where $cp_i$ is
the power it requires, and $d_i$ is the demand, which designates
how many cards of that type have to be plugged. In order to plug a
card into a rack, the rack needs to be assigned a rack model.

Each rack model $i$ in the set $\mydomain{RackModels}$ is
characterised by a tuple $\la rp_i, c_i, s_i \ra$, where $rp_i$ is
the maximal power it can supply, $c_i$ is its number of
connectors, and $s_i$ is its price. Each card plugged into a rack
uses a connector. The problem is to decide how many among the set
$\mydomain{Racks}$ of available racks are needed, and which model
the racks are in order to plug all the cards such that:
\begin{itemize}
\item
the number of cards plugged into a rack does not exceed its number
of connectors;
\item
the total power of the cards plugged into a rack does not exceed
its power;
\item
all the cards are plugged into some rack;
\item
the total price of the racks is minimised.
\end{itemize}

\begin{figure}[t!]
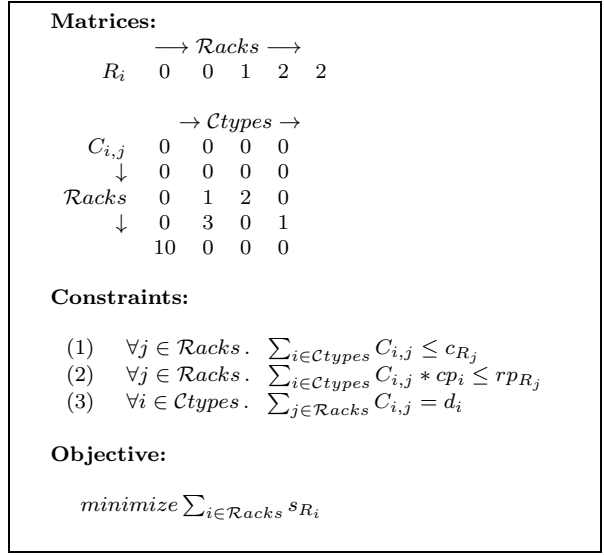

\begin{center}
\framebox{
\begin{scriptsize}
\begin{tabular}{l}
\mbox{{\bf Matrices:}} \\
$
\begin{array}{rccccccccc}
& \multicolumn{6}{c}{\mbox{$\longrightarrow \mydomain{Racks} \longrightarrow$}} \\
R_i        &0&0&1&2&2 \\
        & & & & & \\
& \multicolumn{5}{c}{\mbox{$\rightarrow \mydomain{Ctypes} \rightarrow$}} \\
C_{i,j}         &0&0&0&0\\
\downarrow      &0&0&0&0\\
\mydomain{Racks}&0&1&2&0\\
\downarrow      &0&3&0&1\\
                &10&0&0&0\\
\end{array}$ \\\\
\mbox{{\bf Constraints:}} \\\\
\begin{tabular}{ll}
(1) & $\forall j \in \mydomain{Racks}\Dot~\sum_{i \in \mydomain{Ctypes}} C_{i,j} \leq c_{R_j} $\\
(2) & $\forall j \in \mydomain{Racks}\Dot~\sum_{i \in \mydomain{Ctypes}} C_{i,j}* cp_i \leq rp_{R_j}$ \\
(3) & $\forall i \in \mydomain{Ctypes}\Dot~\sum_{j \in
\mydomain{Racks}} C_{i,j} = d_i$
\end{tabular}\\\\
\mbox{{\bf Objective:}}\\\\
$~~~~minimize \sum_{i \in \mydomain{Racks}} s_{R_i}$ \\\\
\end{tabular}
\end{scriptsize}
}
\end{center}
\caption{The matrix model of the rack configuration problem in
\cite{IlogSolver}.} \label{ch-matrixmodels:table-rack1}
\end{figure}

A matrix model of this problem is given in \cite{IlogSolver} and
shown in Figure \ref{ch-matrixmodels:table-rack1}. The idea is to
assign a rack model to every available rack. Since some of the
racks might not be needed in an optimal solution, a ``dummy'' rack
model is introduced (i.e., a rack is assigned the dummy rack model
when the rack is not needed). Furthermore, for every available
rack, the number of cards of a particular card type
plugged into the rack has to be determined. The assignment of rack
models to racks is represented by a 1-d matrix $R$, indexed by
$\mydomain{Racks}$, taking values from $\mydomain{RackModels}$
which includes the dummy rack model. In order to represent the
number of cards of a particular card type plugged into a
particular rack, a 2-d matrix $C$ of $\mydomain{Ctypes} \times
\mydomain{Racks}$ is introduced. A variable in this matrix takes
values from $\{0,\ldots,maxConn\}$ where $maxConn$ is the maximum
number of cards that can be plugged into any rack.

The dummy rack model is defined as a rack model where the maximal
power it can supply, its number of connectors, and its price are
all set to $0$. The constraints enforce that the connector and the
power capacity of each rack is not exceeded and every card type
meets its demand. The objective is then to minimise the total cost
of the racks.

The 2-d matrix $C$ has partial row symmetry, because racks of the same
rack model are indistinguishable and therefore their card
assignments can be interchanged.
To break this symmetry, we post multiset ordering constraints on
the rows conditionally. Given two racks $i$ and $j$, we enforce
that the rows corresponding to such racks are multiset ordered if
the racks are assigned the same rack model. That is:
\[
R_i=R_j \Implies \la C_{0,i}, \ldots, C_{n-1,i} \ra \leq_m \la
C_{0,j}, \ldots, C_{n-1,j} \ra
\]
where $n$ is the number of card types. We impose such constraints
by either using our filtering algorithm {\MsetLeq} or the
arithmetic constraint. Unfortunately, we are unable to compare
{\MsetLeq} against the $gcc$ decomposition in this problem, as
Solver 5.3 does not allow us to post {\bf IloDistribute}
constraint conditionally.

\begin{table}[t]
\begin{footnotesize}
\begin{center}
\begin{tabular}{||c|c|c|c||}
\hline
Rack Model & Power & Connectors & Price \\
\hline
1 & 150 & 8 & 150 \\
\hline
2 & 200 & 16 & 200 \\
\hline
\end{tabular}
\\

\begin{tabular}{||c|c||}
\hline
Card Type & Power \\
\hline
1 & 20 \\
\hline
2 & 40 \\
\hline
3 & 50 \\
\hline
4 & 75 \\
\hline
\end{tabular}
\caption{\label{ch:gacmset:datarack1} Rack model and card type
specifications in the rack configuration problem
\cite{IlogSolver}.}
\end{center}
\end{footnotesize}
\end{table}

\begin{table}[t!]
\begin{footnotesize}
\begin{center}
\begin{tabular}{||c|cccc||}
\hline Instance & \multicolumn{4}{|c||} {Demand} \\
$\#$     & Type 1 & Type 2 & Type 3 & Type 4\\
\hline 1 & 10 &  4 & 2 & 2  \\
\hline 2 & 10 &  4 & 2 & 4 \\
\hline 3 & 10 & 6 & 2 & 2 \\
\hline 4 & 10 & 4 & 4 & 2 \\
\hline 5 & 10 & 6 & 4 & 2 \\
\hline 6 & 10 & 4 & 2 & 4 \\
\hline
\end{tabular}
\end{center}
\caption{\label{ch:gacmset:datarack2} Demand specification for the
cards in the rack configuration problem.}
\end{footnotesize}
\end{table}

We consider several instances of the rack configuration problem,
which are described in Tables \ref{ch:gacmset:datarack1} and
\ref{ch:gacmset:datarack2}. In the experiments, we use the rack
model and card type specifications given in \cite{IlogSolver}, but
we vary the demand of the card types randomly. As in
\cite{IlogSolver}, we search for the optimal solution by exploring
the racks in turn. For each rack, we first instantiate its model
and then determine how many cards from each card type are plugged
into the rack.

\begin{table}[t!]
\begin{footnotesize}
\begin{center}
\begin{tabular}{||c||cccc||}
\hline          
Inst. & \multicolumn{3}{c}{{\MsetLeq} R}& {Arithmetic Constraint R} \\
        $\#$ & Fails & Choice & Time & Time  \\
        &       & points & (secs.)& (secs.) \\
\hline
1 &  3,052 & 3,063 & {\bf 0.2} & 2.8  \\
\hline
2 & 15,650 & 15,657 & {\bf 0.6}  & 15.6 \\
\hline
3 & 3,990 &  3,999 & {\bf 0.2}& 2.6 \\
\hline
4 & 8,864 & 8,872 & {\bf 0.4} & 7.1 \\
\hline
5 & 40,851 & 40,858 & {\bf 1.5} & 41.3  \\
\hline
6 & 42,013 & 42,026 & {\bf 1.6} & 35.2  \\

\hline
\end{tabular}
\end{center}
\caption{\label{ch:gacmset:RackResults} Rack configuration
problem: {\MsetLeq} vs the arithmetic constraint.}
\end{footnotesize}
\end{table}

The results of the experiments are shown in Table
\ref{ch:gacmset:RackResults}. {\MsetLeq} is clearly much more
efficient than the arithmetic constraint on every instance
considered. Note that the two methods create the same search tree.

\subsection{Sport Scheduling Problem}

This problem was introduced in Section \ref{sec:back}. Figure
\ref{ch-matrixmodels:table-sport} shows a matrix model.
The (extended) weeks over which the tournament
is held, as well the periods of a week are indistinguishable. The
rows and the columns of $T$ and $G$ are therefore symmetric. Note
that we treat $T$ as a 2-d matrix where the rows represent the
periods and columns represent the (extended) weeks, and each entry
of the matrix is a pair of values.
The global cardinality constraints posted on the rows of $T$ ensure
that each of $1 \ldots n$ occur exactly twice in every row.
In any solution to the problem, the rows when
viewed as multisets are therefore equal.
The $\myalldiff$
constraints posted on the columns state that each column is a
permutation of $1 \ldots n$. Thus,
the columns are also equal when viewed as multisets.
Therefore, we cannot utilise multiset ordering constraints to break
row and/or column symmetry of this model of the problem.

Scheduling a tournament between $n$ teams means arranging $n(n-1)/2$
games. The model described in Figure
\ref{ch-matrixmodels:table-sport} assumes $n$ is an even number. If
$n$ is an odd number instead, then we can still schedule $n(n-1)/2$
games provided that the games are played over $n$ weeks and each
week is divided into $(n-1)/2$ periods. The problem now requires
that each team plays at most once a week, and every team plays
exactly twice in the same period over the tournament. This version
of the problem can be modelled using the original model in Figure
\ref{ch-matrixmodels:table-sport}, as the {\em all-different}
constraints on the rows and the cardinality constraints on the
columns enforce the new problem constraints.

\begin{table}[t]
\begin{footnotesize}

\begin{center}
\begin{tabular}{||c||l||r|r|l||}
\hline $n$ &  Model & Fails & Choice points & Time (sec.) \\
\hline
$5$  &{\MsetLess} C& {\bf 1} & {\bf 10} & {\bf 0.8}  \\
& Arithmetic Constraint C &  1& 10& 0.9 \\
& $gcc$ C & 2 & 11& 1.2  \\


\hline

$7$ & {\MsetLess} C & {\bf 69} & {\bf 87} & {\bf 0.8}  \\
& Arithmetic Constraint C &  69 & 87 & 1.3 \\
& $gcc$ C & 74 & 92 & 1.3  \\


\hline

$9$ &{\MsetLess} C & 760,973 & 761,003 & {\bf 121.3}  \\
&Arithmetic Constraint C & 760,973 & 761,003 & 2500 \\
&$gcc$ C & 2,616,148 & 2,616,176 & 656.4  \\


\hline

\end{tabular}
\end{center} \caption{\label{ch:gacmset:OddSportsResults-1} Sport scheduling problem: {\MsetLess}
vs $gcc$ decomposition and the arithmetic constraint  with
column-wise labelling. For one column, we first label the first
slots; for the other, we first label the second slots. }
\end{footnotesize}

\end{table}

We can now post multiset ordering constraints on the columns of $T$
to break column symmetry. Since the games are all different, no pair
of columns can be equal, when viewed as multisets. Hence, we insist
that the columns corresponding to the $n$ weeks
are strict multiset ordered: $\myvec{C_0}<_{m} \myvec{C_1} \ldots
<_{m} \myvec{C_{n-1}}$. We enforce such constraints by either using
our filtering algorithm {\MsetLess}, or the $gcc$ decomposition, or
the arithmetic constraint. Since the multiset ordering constraints
are posted on the columns, we instantiate $T$ column-by-column.
For one column, we first label
the first slots; for the other, we first label the second slots. The
results are shown in Table \ref{ch:gacmset:OddSportsResults-1}.


We observe that {\MsetLess} is superior to the $gcc$ decomposition.
As the problem gets more difficult, {\MsetLess} does more pruning
and solves the problem quicker.
The results moreover indicate a substantial gain in efficiency by
using {\MsetLess} in preference to the arithmetic constraint. Even
though the same search tree is created by the two, constructing and
propagating the arithmetic constraints is much more costly than
running {\MsetLess} to solve the multiset ordering constraints.

\section{Conclusions}
\label{ch-gacmset:summary}

We have developed filtering algorithms for the multiset ordering
(global) constraint $\myvec{X} \leq_m \myvec{Y}$ defined on a pair
of vectors of variables. It ensures that the values taken by the
vectors $\myvec{X}$ and $\myvec{Y}$, when viewed as multisets, are
ordered. This global constraint is useful for breaking row and
column symmetries of a matrix model and when searching for leximin
solutions in fuzzy constraints.
The filtering algorithms either prove that $\myvec{X} \leq_m
\myvec{Y}$ is disentailed, or ensure GAC on $\myvec{X} \leq_m
\myvec{Y}$.

The first algorithm {\MsetLeq} is useful when $d \ll  n$ and runs in
$O(n)$ where $n$ is the length of the vectors and $d$ is the number
of distinct values. This is often the case as the number of distinct
values in a multiset is typically less than its cardinality to
permit repetition. We further proposed another variant of the
algorithm suitable when $d \gg n$. This identifies support by
lexicographically ordering suitable sorted vectors. The complexity
is then independent of the number of distinct
values and is $O(n~log(n))$, as the cost of
sorting dominates. We also have shown that  {\MsetLeq} can easily be
modified for $\myvec{X} <_{m} \myvec{Y}$ by changing the definition
of one of the flags. Moreover, the ease of maintaining the
occurrence vectors incrementally helps detect entailment in a simple
and dual manner to detecting disentailment.



Our experiments on the the progressive party problem, the rack
configuration problem, and and the sport scheduling problem support
the usefulness of multiset ordering constraints in the context of
symmetry breaking and support our theoretical studies: even if it is
feasible to post the arithmetic constraint, it is much more
efficient to propagate the multiset ordering constraint using our
filtering algorithm; furthermore, decomposing the multiset
constraint carries penalty either in the amount or the cost of
constraint propagation.

In our future work, we plan to investigate whether the
incremental cost for propagation can be made
less than linear time. Moreover, we plan to
understand whether it is worthwhile to a propate a chain of multiset
ordering constraints and if that is the case devise an efficient
filtering algorithm.



\section*{Acknowledgements} The authors would like to thank the
anonymous reviewers for their useful comments on the presentation
and Chris Jefferson for fruitful discussions on the work described
in the article. B. Hnich is supported by Scientific and
Technological Research Council of Turkey (TUBITAK) under Grant No:
SOBAG-108K027. I. Miguel is supported by a UK Royal Academy of
Engineering/EPSRC Research Fellowship.


\newcommand{\etalchar}[1]{$^{#1}$}

\end{document}

%% file: macros.tex
\newcommand{\la}{\langle}
\newcommand{\ra}{\rangle}

\newcommand{\Dot}{\,{\bf .}\;}  
\newcommand{\Or}{\vee}
\newcommand{\And}{\wedge}
\newcommand{\Not}{\neg}
\newcommand{\Implies}{\rightarrow}
\newcommand{\Iff}{\leftrightarrow}

\newenvironment{Rem}[1]{\leavebigvspace\noindent
{\tt ** #1 **}\newline}{$\phantom{a}$\newline{\tt ********}\leavebigvspace}

\def\halmos{\mbox{\hspace*{1pc}$\Box$}}

\newcommand{\leavebigvspace}{\par \addvspace{\bigskipamount}}
\def\theexample{\thesection.\arabic{example}}
\newtheorem{Example}{Example}[section]
{{$\phantom{a}$%
\leavebigvspace}%
\end{Example}}
\def\thetheorem{\thesection.\arabic{theorem}}
\newtheorem{Theorem}{Theorem}[section]
{{$\phantom{a}$}\end{Theorem}}
\def\thedefinition{\thesection.\arabic{definition}}
\newtheorem{Definition}{Definition}[section]
{{$\phantom{a}$}\end{Definition}}